\documentclass[conference]{IEEEtran}
\usepackage{times}

\usepackage[numbers,sort&compress]{natbib}
\usepackage{multicol}
\usepackage[bookmarks=true]{hyperref}

\usepackage[utf8]{inputenc} %
\usepackage[T1]{fontenc}    %
\usepackage{url}            %
\usepackage{amsfonts}       %
\usepackage{nicefrac}       %
\usepackage{microtype}      %
\usepackage{tabularx}
\usepackage{microtype}
\usepackage{graphicx}
\usepackage{subcaption}
\usepackage{hyperref}
\hypersetup{
	colorlinks=true,
	linkcolor=orange,
	filecolor=magenta,      
	urlcolor=orange,
	citecolor=orange,
}
\usepackage[utf8]{inputenc} %
\usepackage[T1]{fontenc}    %
\usepackage{dsfont}
\usepackage{url}            %
\usepackage{amsfonts}       %
\usepackage{nicefrac}       %
\usepackage{color}
\usepackage[dvipsnames]{xcolor}
\usepackage{mathtools}
\usepackage{amsmath,amssymb}
\usepackage{bm}
\usepackage{siunitx}
\usepackage{wrapfig}
\usepackage{algorithm}
\usepackage[noend]{algorithmic}
\usepackage{lipsum}
\usepackage{makecell}
\usepackage{cite}
\usepackage{subcaption}
\usepackage[capitalise, nameinlink]{cleveref}

\usepackage{footnote}
\makesavenoteenv{tabular}
\makesavenoteenv{table}
\usepackage{multirow}
\usepackage{booktabs}
\usepackage{amsmath}
\usepackage{amssymb}
\usepackage{mathtools}

\usepackage{pifont}%
\newcommand{\cmark}{\ding{51}}%
\newcommand{\xmark}{\ding{55}}%

\usepackage{amsmath,amsfonts,bm}

\def\eqref#1{equation~\ref{#1}}

\def\1{\bm{1}}

\DeclareMathAlphabet{\mathsfit}{\encodingdefault}{\sfdefault}{m}{sl}
\SetMathAlphabet{\mathsfit}{bold}{\encodingdefault}{\sfdefault}{bx}{n}

\DeclarePairedDelimiterX{\norm}[1]{\lVert}{\rVert}{#1}

\newcommand{\ie}{i.e., }
\newcommand{\eg}{e.g., }
\newcommand{\Skip}[1]{}

\usepackage{verbatim}
\usepackage{mdframed}
\mdfdefinestyle{codebox}{%
    backgroundcolor=gray!25, %
    linewidth=0pt, %
}
\usepackage{newfloat}
\DeclareFloatingEnvironment[
    fileext=los,
    listname={List of Listings},
    name=Listing,
    placement=tbhp,
    within=section,
]{listing}

\newcommand{\name}{DROID}
\newcommand{\namelong}{DROID (\textbf{D}istributed \textbf{Ro}bot \textbf{I}nteraction \textbf{D}ataset)}
\newcommand{\ntrajs}{76k}
\newcommand{\nfailtrajs}{16k}
\newcommand{\ninstitutions}{13}
\newcommand{\nlabs}{18}
\newcommand{\nscenes}{564}
\newcommand{\nhours}{350}
\newcommand{\nbuildings}{52}
\newcommand{\ntasks}{86}
\newcommand{\nmonths}{12}
\newcommand{\ncollectors}{50}
\newcommand{\nrobots}{18}
\newcommand{\nevaltasks}{6}
\newcommand{\nevallocations}{4}
\newcommand{\ncamposes}{1417}
\newcommand{\license}{CC-BY 4.0}
\newcommand{\website}{\url{https://droid-dataset.github.io}}

\usepackage[font=small]{caption} %
\begin{document}

\title{\vspace{-0.5cm}\name{}: A Large-Scale In-The-Wild \\Robot Manipulation Dataset\\[0.3cm]\large\website
\vspace{-0.5cm}}

\author{Alexander Khazatsky$^{\ast, 1}$, Karl Pertsch$^{\ast, 1, 2}$, Suraj Nair$^{1, 3}$, Ashwin Balakrishna$^{3}$, Sudeep Dasari$^{4}$, \\
Siddharth Karamcheti$^{1}$, Soroush Nasiriany$^{5}$, Mohan Kumar Srirama$^{4}$, Lawrence Yunliang Chen$^{2}$, Kirsty Ellis$^{6}$, \\
Peter David Fagan$^{7}$, Joey Hejna$^{1}$, Masha Itkina$^{3}$, Marion Lepert$^{1}$, Jason Ma$^{14}$, Patrick Tree Miller$^{3}$, \\Jimmy Wu$^{8}$, Suneel Belkhale$^{1}$, Shivin Dass$^{5}$, Huy Ha$^{1}$, Abraham Lee$^{2}$, Youngwoon Lee$^{2,16}$, Arhan Jain$^{9}$, \\
Marius Memmel$^{9}$, Sungjae Park$^{10}$, Ilija Radosavovic$^{2}$, Kaiyuan Wang$^{11}$, Albert Zhan$^{6}$, Kevin Black$^{2}$, \\
Cheng Chi$^{1}$, Kyle Hatch$^{3}$, Shan Lin$^{11}$, Jingpei Lu$^{11}$, Abdul Rehman$^{7}$, Pannag R Sanketi$^{12}$, \\
Archit Sharma$^{1}$, Cody Simpson$^{3}$, Quan Vuong$^{12}$, Homer Walke$^{2}$, Blake Wulfe$^{3}$, Ted Xiao$^{12}$, Jonathan Yang$^{1}$, \\
Arefeh Yavary$^{13}$, Tony Z. Zhao$^{1}$, Christopher Agia$^{1}$, Rohan Baijal$^{9}$, Mateo Guaman Castro$^{9}$, Daphne Chen$^{9}$, \\
Qiuyu Chen$^{9}$, Trinity Chung$^{2}$, Jaimyn Drake$^{2}$, Ethan Paul Foster$^{1}$, Jensen Gao$^{1}$, Vitor Guizilini$^{3}$, \\
David Antonio Herrera$^{1}$, Minho Heo$^{10}$, Kyle Hsu$^{1}$, Jiaheng Hu$^{5}$, Muhammad Zubair Irshad$^{3}$, Donovon Jackson$^{3}$, \\
Charlotte Le$^{2}$, Yunshuang Li$^{14}$, Kevin Lin$^{1}$, Roy Lin$^{2}$, Zehan Ma$^{2}$, Abhiram Maddukuri$^{5}$, Suvir Mirchandani$^{1}$,\\
Daniel Morton$^{1}$, Tony Nguyen$^{3}$, Abby O'Neill$^{2}$, Rosario Scalise$^{9}$, Derick Seale$^{3}$, Victor Son$^{1}$, Stephen Tian$^{1}$, \\
Andrew Wang$^{2}$, Yilin Wu$^{4}$, Annie Xie$^{1}$, Jingyun Yang$^{1}$, Patrick Yin$^{9}$, Yunchu Zhang$^{9}$, \\
Osbert Bastani$^{14}$, Glen Berseth$^{6}$, Jeannette Bohg$^{1}$, Ken Goldberg$^{2}$, Abhinav Gupta$^{4}$, Abhishek Gupta$^{9}$, \\
Dinesh Jayaraman$^{14}$, Joseph J. Lim$^{10}$, Jitendra Malik$^{2}$, Roberto Martín-Martín$^{5}$, Subramanian Ramamoorthy$^{7}$, \\
Dorsa Sadigh$^{1}$, Shuran Song$^{1, 15}$, Jiajun Wu$^{1}$, Yuke Zhu$^{5}$, Thomas Kollar$^{3}$, Sergey Levine$^{2}$, Chelsea Finn$^{1}$
}

\makeatletter
\let\@oldmaketitle\@maketitle%
\renewcommand{\@maketitle}{\@oldmaketitle%
  \begin{center}
  \captionsetup{type=figure}
  \includegraphics[width=\textwidth]{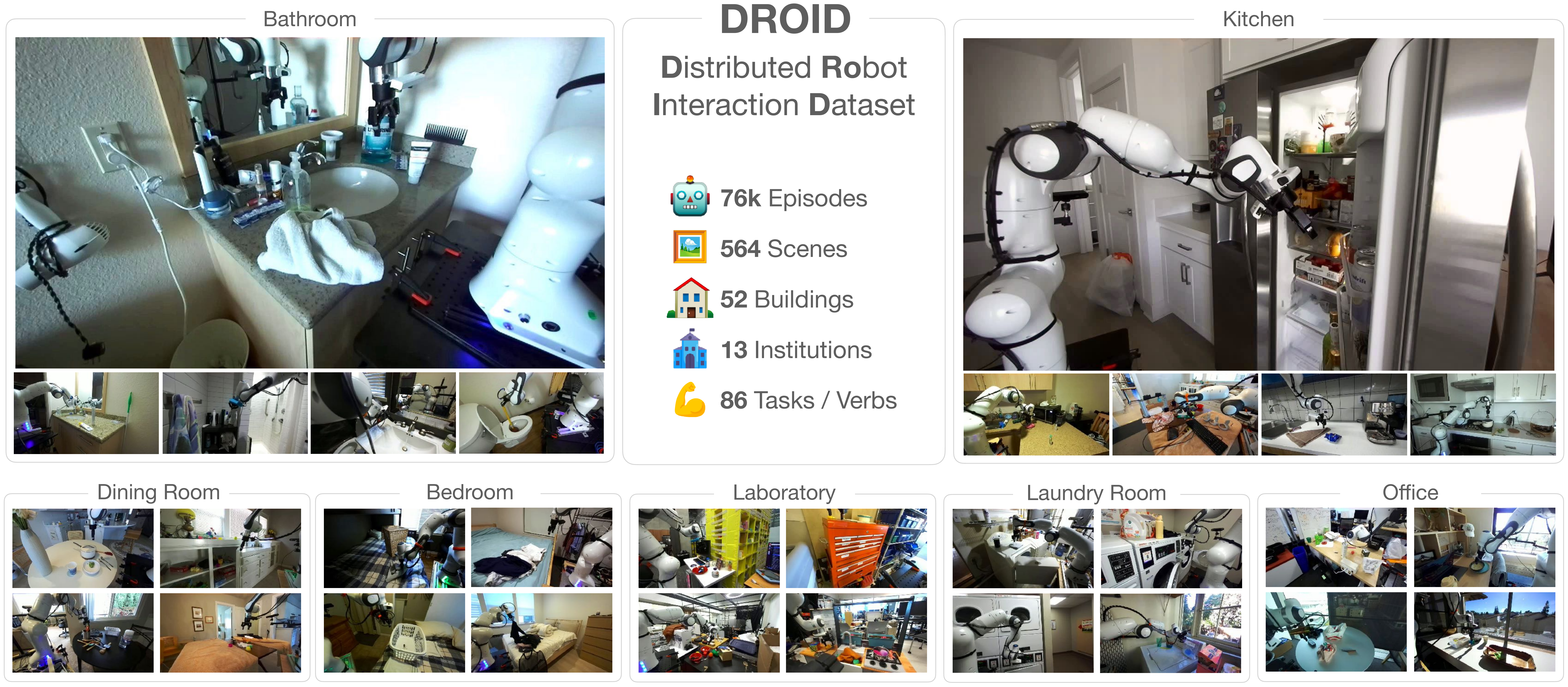}
    \captionof{figure}{We introduce \namelong{}, an ``in-the-wild'' robot manipulation dataset with \ntrajs{}~trajectories or \nhours{}~hours of interaction data, collected across \nscenes{}~scenes, \ntasks{}~tasks, and \nbuildings{}~buildings over the course of \nmonths~months. Each \name{} episode contains three synchronized RGB camera streams, camera calibration, depth information, and natural language instructions. We demonstrate that training with \name{} leads to policies with higher performance, greater robustness, and improved generalization ability. We open source the full dataset, pre-trained model checkpoints, and a detailed guide for reproducing our robot setup.} 
    \label{fig:teaser}
    \end{center}
}
\makeatother

\maketitle
\addtocounter{figure}{-1}

\renewcommand*{\thefootnote}{\arabic{footnote}}

\begin{abstract}
\renewcommand*{\thefootnote}{\fnsymbol{footnote}}
\footnotetext{$^\ast$Project Co-leads, correspondence to \href{mailto:alexkhaz@stanford.edu,pertsch@berkeley.edu}{\texttt{alexkhaz@stanford.edu, pertsch@berkeley.edu}}}
\renewcommand*{\thefootnote}{\arabic{footnote}}
The creation of large, diverse, high-quality robot manipulation datasets is an important stepping stone on the path toward more capable and robust robotic manipulation policies. However, creating such datasets is challenging: collecting robot manipulation data in diverse environments poses logistical and safety challenges and requires substantial investments in hardware and human labour. As a result, even the most general robot manipulation policies today are mostly trained on data collected in a small number of environments with limited scene and task diversity. 
In this work, we introduce \namelong{}, a diverse robot manipulation dataset with \ntrajs{}~demonstration trajectories or \nhours{}~hours of interaction data, collected across \nscenes{}~scenes and \ntasks{}~tasks by \ncollectors{}~data collectors in North America, Asia, and Europe over the course of \nmonths{}~months. We demonstrate that training with \name{} leads to policies with higher performance and improved generalization ability. 
We open source the full dataset, policy learning code, and a detailed guide for reproducing our robot hardware setup.

\end{abstract}

\section{Introduction}
\label{sec:intro}
\renewcommand*{\thefootnote}{\fnsymbol{footnote}}
\footnotetext{Affiliations: $^{1}$Stanford University; $^{2}$University of California, Berkeley; $^{3}$Toyota Research Institute; $^{4}$Carnegie Mellon University; $^{5}$University of Texas, Austin; $^{6}$University of Montreal; $^{7}$University of Edinburgh; $^{8}$Princeton University; $^{9}$University of Washington; $^{10}$Korea Advanced Institute of Science \& Technology (KAIST); $^{11}$University of California, San Diego; $^{12}$Google DeepMind; $^{13}$University of California, Davis; $^{14}$University of Pennsylvania; $^{15}$Columbia University; $^{16}$Yonsei University
}
\renewcommand*{\thefootnote}{\arabic{footnote}}

\begin{table*}[t]
    \centering
    \setcounter{footnote}{1}
        \begin{tabular}{l c c c c c c l}
            \toprule
            Dataset     & \# Traj.  & \# Verbs & \# Scenes & Lang. Instruct.  &  Cam. Calibration  & Public Robot & Collection \\
            \midrule
            MIME~\citep{sharma2018multiple}        & 8.3k     & 20     & 1       & \xmark & \xmark     & \cmark       & human teleop                                   \\
            RoboTurk~\citep{mandlekar2018roboturk}    & 2.1k     & 2      & 1       & \xmark & \xmark     & \cmark       & human teleop                                    \\
            RoboNet~\citep{dasari2019robonet}     & 162k      & n/a    & 10      & \xmark & \xmark   & \cmark       & scripted                                 \\
            MT-Opt~\citep{kalashnikov2018qt,mtopt2021arxiv}      & 800k      & 2      & 1       & \xmark & \xmark   & \cmark       & \mbox{scripted \& learned}               \\
            BridgeData~\citep{ebert2021bridge}  & 7.2k     & 4      & 12      & \cmark & \xmark     & \cmark       & human teleop                                    \\
            BC-Z~\citep{jang2022bc}        & 26k     & 3      & 1       & \cmark & \xmark   & \xmark       & human teleop                                    \\
            RT-1~\citep{brohan2022rt}        & 130k      & 2      & 2       & \cmark & \xmark   & \xmark       & human teleop                                    \\
            RH20T~\citep{fang2023rh20t}       & 13k\footnotemark      & 33     & 7   & \cmark & \cmark    & \cmark       & human teleop                                    \\
            RoboSet~\citep{bharadhwaj2023roboagent}     & 98.5k     & 9      & 11      & \cmark & \xmark    & \cmark       & 30\% human / 70\%  scripted      \\
            BridgeData~V2~\citep{walke2023bridgedata}  & 60.1k  & 82     & 24      & \cmark & \xmark    & \cmark       & 85\% human / 15\%  scripted \\[0.4em]
            \textcolor{gray}{DobbE~\citep{shafiullah2023dobbe}$^*$}  & \textcolor{gray}{5.6k}  & \textcolor{gray}{6}     & \textcolor{gray}{216}      & \textcolor{gray}{\cmark} & \textcolor{gray}{n/a}    & \textcolor{gray}{(\cmark)}       & \textcolor{gray}{human tool-based} \\
            \textcolor{gray}{Open~X-Embodiment~\citep{open_x_embodiment_rt_x_2023}$^\dagger$}   & \textcolor{gray}{1.4M}      & \textcolor{gray}{217}     & \textcolor{gray}{311}   & \textcolor{gray}{(\cmark)} & \textcolor{gray}{\xmark}     & \textcolor{gray}{(\cmark)}       & \textcolor{gray}{dataset aggregation}  \\[0.2em]
            \midrule
            \textbf{\name{} (ours)} & \ntrajs{} & \ntasks{} & \textbf{\textcolor{red}{\nscenes{}}} & \cmark & \cmark & \cmark & human teleop \\
            \bottomrule                                                                                                                                             \\
        \end{tabular}
    \setcounter{footnote}{0}
    \captionsetup{width=\textwidth}
    \caption{Comparison to existing datasets for robot manipulation. ``\# Scenes'' refers to the number of unique robot work spaces, \eg different kitchens count as different scenes, but rearrangement of objects does not. See \cref{sec:related_work} for a detailed discussion of the definition of ``Tasks'' and ``Scenes''.  
    \name{} offers high diversity in both, the number of verbs and scenes. \small{\textcolor{gray}{$^*$non-robot, tool-based data collection, $^\dagger$not a dataset in itself, but \emph{aggregation} of existing datasets, including most previous rows in this table.}}
    } 
    \label{tab:dataset_comp}
\end{table*}

A key feature of robot manipulation policies is their ability to generalize, \ie their ability to perform a desired manipulation task under new lighting conditions, in new environments, or with new objects. Training policies that are robust to such variations is a crucial step towards the deployment of robots in everyday environments and may bring us closer to every roboticist's dream: robot models that can be downloaded and ``just work'' when tested on a new robot setup. A central ingredient for training such generalizable policies is diverse training data: in computer vision and natural language processing, training on large and diverse datasets scraped from the internet yields models that work in a wide range of new tasks. 
Similarly, in robot manipulation, a number of recent works have demonstrated that larger, more diverse robot training datasets enable us to push the envelope on policy generalization, including positive transfer to new objects, instructions, scenes, and embodiments~\citep{bharadhwaj2023roboagent, brohan2022rt, fang2023rh20t,mandlekar2018roboturk, octo_2023, open_x_embodiment_rt_x_2023,shafiullah2023dobbe,young2020visual}.
This suggests that an important stepping stone on the path toward more capable and robust robotic manipulation policies is the creation of large, diverse, high-quality robot manipulation datasets. 

However, creating such datasets is challenging: in contrast to vision or language data, training manipulation policies typically requires robot manipulation data with recorded observations and actions, which cannot be easily scraped from the internet. Collecting robot manipulation data in diverse environments poses logistical and safety challenges when moving robots outside of controlled lab environments. Additionally, collecting data at scale requires substantial investments in hardware and human labour for supervision, particularly for collecting demonstration data. 
As a result, even the most general robot manipulation policies today are mostly trained on data collected in controlled, lab-like environments with limited scene and task diversity. To enable the next level of generalizable robot manipulation policy learning, the robot manipulation community needs more diverse datasets, collected across a wide range of environments and tasks.

In this work, we introduce \namelong{},
a robot manipulation dataset of unprecedented diversity (see \cref{fig:teaser}). \name{} consist of \ntrajs{}~demonstration trajectories or \nhours{}~hours of interaction data, collected across \nscenes{}~scenes, \nbuildings{}~buildings and \ntasks{}~tasks. \name{} was collected by \nlabs{}~research labs
in North America, Asia, and Europe 
over the course of \nmonths{}~months. 
To streamline distributed data collection and ensure applicability of the final dataset to a wide range of research settings, all data is collected on the same robot hardware stack based on the popular Franka Panda robot arm. Each episode contains three camera views, depth information, camera calibration, and language annotations. 

In experiments across \nevaltasks{}~tasks and \nevallocations{}~locations, from labs to offices and real households, we find that \name{} boosts policy performance, robustness and generalizability by 20\% on average over state-of-the-art approaches that leverage existing large-scale robot manipulation datasets~\citep{open_x_embodiment_rt_x_2023}. 
We open-source the full \name{}~dataset under \license{}~license, code for training policies using the dataset, and a detailed guide for reproducing our complete robot software and hardware setup. %

\section{Related Work}
\label{sec:related_work}

\paragraph{Large datasets in machine learning}
The rapid progress in machine learning has been closely tied to the construction of large and diverse datasets. Examples include ImageNet~\citep{deng2009imagenet}, Kitti~\citep{geiger2012we}, Ego4D~\citep{grauman2022ego4d} and LAION~\citep{schuhmann2022laion} in computer vision, Common Crawl~\citep{cc:Rana:2010:Common-Crawl-open-web-scale-crawl} and The~Pile~\citep{gao2020pile} in natural language processing, and ShapeNet~\citep{chang2015shapenet} and Objaverse~\citep{deitke2023objaverse,deitke2023objaversexl} in 3D modeling. Key to their impact is their size and diversity: by enabling training on larger and more diverse data, they push the capabilities and robustness of machine learning models. With \name{} we aim to continue this trend for robot manipulation and provide a large and diverse robot manipulation dataset to spur progress on generalizable policy learning.

\paragraph{Robot learning datasets}
A number of prior works introduce datasets for robot learning of various sizes and diversity levels (see \cref{tab:dataset_comp}). Broadly, these can be categorized into datasets collected autonomously via scripted and semi-random behaviors or learned agents~\citep{levine2016learning,pinto2016supersizing,kalashnikov2018qt,gupta2018robot,dasari2019robonet,cabi2019,mtopt2021arxiv}, and datasets collected via human teleoperation~\citep{mandlekar2018roboturk,sharma2018multiple,ebert2021bridge,jang2022bc,brohan2022rt,walke2023bridgedata,fang2023rh20t,bharadhwaj2023roboagent}. Multiple works focus on increasing dataset diversity: RH20T~\citep{fang2023rh20t} collects data across 33 tasks in 7 table-top scenes and BridgeV2~\citep{walke2023bridgedata} collects data in 24 scenes.\footnote{Note that prior works use various definitions for what constitutes a ``task'' and what constitutes a ``scene''. In this work, we use the number of unique \emph{verbs} extracted from the language instructions to represent the number of tasks, which is more scalable than manually defining tasks~\citep{walke2023bridgedata} yet often more reflective of the behavior diversity than \eg counting the number of verb-object combinations~\citep{brohan2022rt} (see \cref{fig:verb_object_distribution} for \name{}'s verb distribution as an example). For scenes, we only count a scene as new if there is a substantial change of the robot's workspace, \eg if it gets transported to a new corner of the kitchen or a new room altogether, but not if only the arrangement of objects in front of the robot or the table cloth changes.} While these datasets increase diversity, most of their data is collected in a small number of scenes in a single research lab or building. 

More recently, there has been a larger effort on pooling existing robot datasets into a coherent format, the Open X-Embodiment dataset (OXE)~\citep{open_x_embodiment_rt_x_2023}. Albeit larger in scale than prior robot datasets, the OXE dataset still consists of individual datasets with few scenes, thus totalling around 300 scenes at the time of writing. Our goal with the \name{} dataset is to significantly increase the scene diversity as well as scene realism by collecting data across a wide array of real world buildings in a diverse set of geographic locations. As a result, \name{} contains data from \nscenes{}~scenes across \nbuildings{}~buildings, a substantial increase compared to any existing robot manipulation dataset.

Collecting such data ``in-the-wild'' is more common for robot navigation and autonomous driving~\citep{geiger2012we,nuscenes2019,sun2019scalability,yu2020bdd100k,karnan2022socially,triest2022tartandrive,shahGNMGeneralNavigation2023,shah2023vint} and enables training of policies that generalize zero-shot to new environments and even embodiments~\citep{shahGNMGeneralNavigation2023,shah2023vint}.
With \name{}, we take a step towards enabling similar generalization for robotic \emph{manipulation} policies.
Finally, there are some works that leverage cheap, off-the-shelf tools, such as reacher-grabber tools, for data collection, equipping robots with the same tools to allow for zero-shot transfer to the robot~\citep{song2020grasping,young2020visual,shafiullah2023dobbe}. 
While this simplifies the data collection process, it limits the data to wrist camera viewpoints and may suffer from morphology differences when transferring from human-arm-collected data to robot arm execution. Additionally, \name{} has larger scene and task diversity than prior tool-based collection datasets~\citep{shafiullah2023dobbe}.

\setcounter{footnote}{2}
\footnotetext{\citet{fang2023rh20t} report 110k trajectories for RH20T, but count each camera stream separately -- here we report the number of unique multi-view trajectories, to compare fairly to all other datasets.}
\setcounter{footnote}{2}

\paragraph{Scalable robot policy learning} Learning robot policies from increasingly large and diverse datasets has been the focus of numerous efforts over the last few years. Initially, these efforts focused in large part on learning from scripted or autonomously collected data~\citep{pinto2016supersizing,levine2016learning,kalashnikov2018qt,dasari2019robonet,ebert2018visual,gupta2018robot}. The success of transformer models~\citep{vaswani2017attention} in natural language processing and computer vision motivated
a number of recent works that collected large-scale demonstration datasets and trained transformer-based policies on them~\citep{brohan2022rt,shridhar2022peract,zitkovich2023rt,open_x_embodiment_rt_x_2023,shah2023vint,radosavovic2023robot,pmlr-v202-jiang23b,zhao2023learning,octo_2023,fu2024mobile}. Additionally, recent works suggest that diffusion denoising models~\citep{ho2020denoising} are a powerful parametrization for multi-modal action output distributions that combine expressivity with scalability~\citep{chi2023diffusionpolicy,ha2023scaling,sridhar2023nomad,octo_2023,fu2024mobile}. Our focus with \name{} is on introducing a new dataset, not a new policy learning algorithm. As such, we build on \emph{existing} state-of-the-art diffusion policies~\citep{chi2023diffusionpolicy} for all of our policy learning experiments.

\section{\name{} Data Collection Setup}
\label{sec:dataset}

In this work, we introduce \namelong{}, an open-source robot manipulation dataset that provides for very high diversity and variability of scenes, tasks, and objects (see \cref{tab:dataset_comp}). Diverse and high-quality data is a key ingredient for training generalizable policies, and \name{} is designed to deliver both quantity and quality: it contains \ntrajs{}~robot demonstration trajectories, spanning \ntasks{}~tasks and \nscenes{}~scenes. It was collected over the course of \nmonths{}~months in a large, cross-institutional effort with \nrobots{}~robots and \ncollectors{}~data collectors across \ninstitutions{}~institutions. All data is collected on a shared, open-source robot platform.

We are releasing all resources to enable researchers to build upon \name{} at \website. This includes the full dataset under \license{} license, an interactive dataset visualizer, code for training generalizable policies on \name{}, pre-trained policy checkpoints, and a detailed guide for reproducing our robot hardware setup and control stack. In this section, we introduce our hardware setup and the data collection protocol.

\subsection{\name{} Robot Platform}
\label{sec:robot_platform}

\begin{figure}[t]
  \centering
  \includegraphics[width=1.0\linewidth]{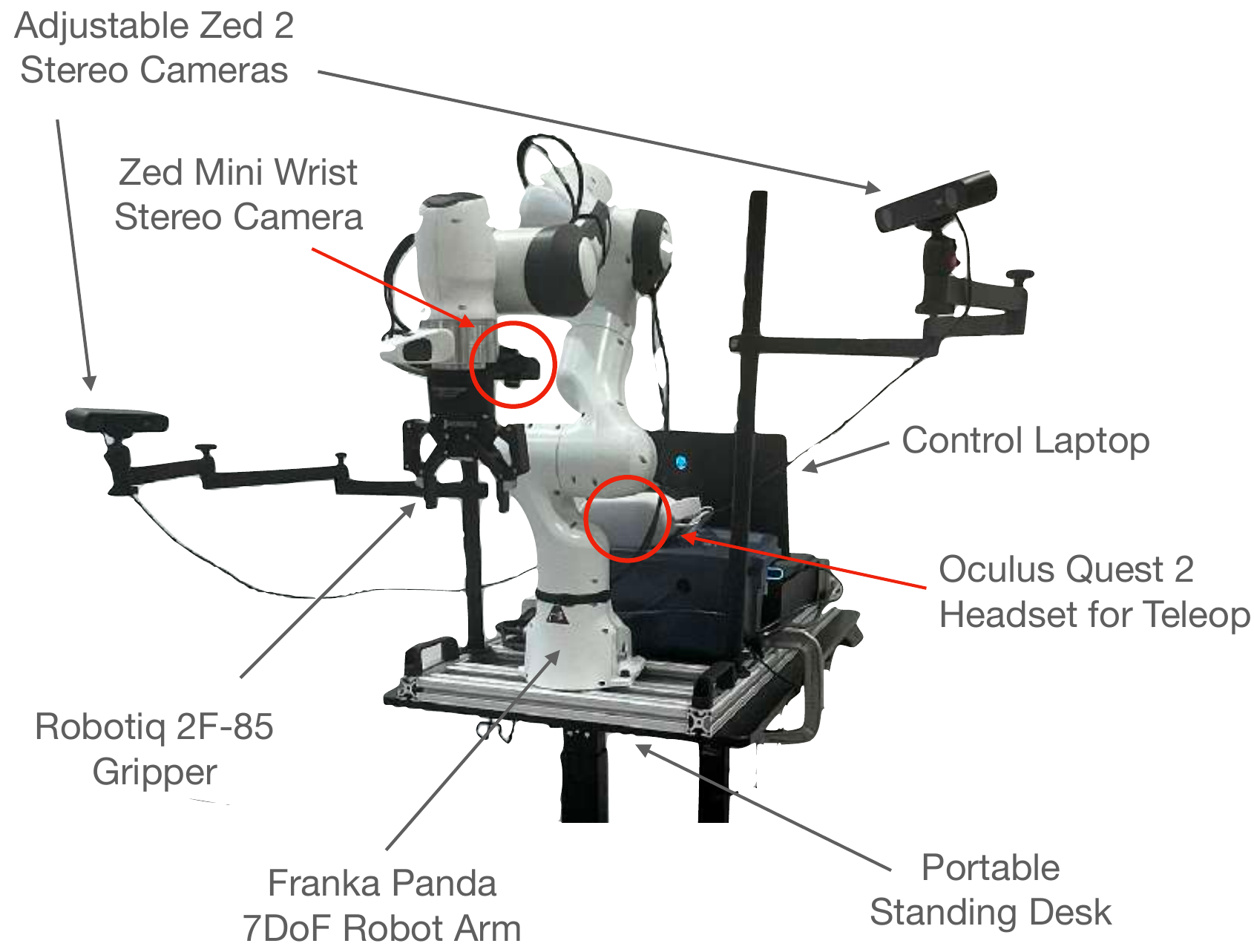}
  \caption{\small The \name{} robot platform. We use the same hardware setup across all \ninstitutions{}~institutions to streamline data collection while maximizing portability and flexibility. The setup consists of a Franka Panda 7DoF robot arm, two adjustable Zed~2 stereo cameras, a wrist-mounted Zed~Mini stereo camera, and an Oculus Quest~2 headset with controllers for teleoperation. Everything is mounted on a portable, height-adjustable desk for quick scene changes.}
  \label{fig:robot_setup}
\end{figure}

\begin{figure*}[t]
  \centering
  \includegraphics[width=1.0\linewidth]{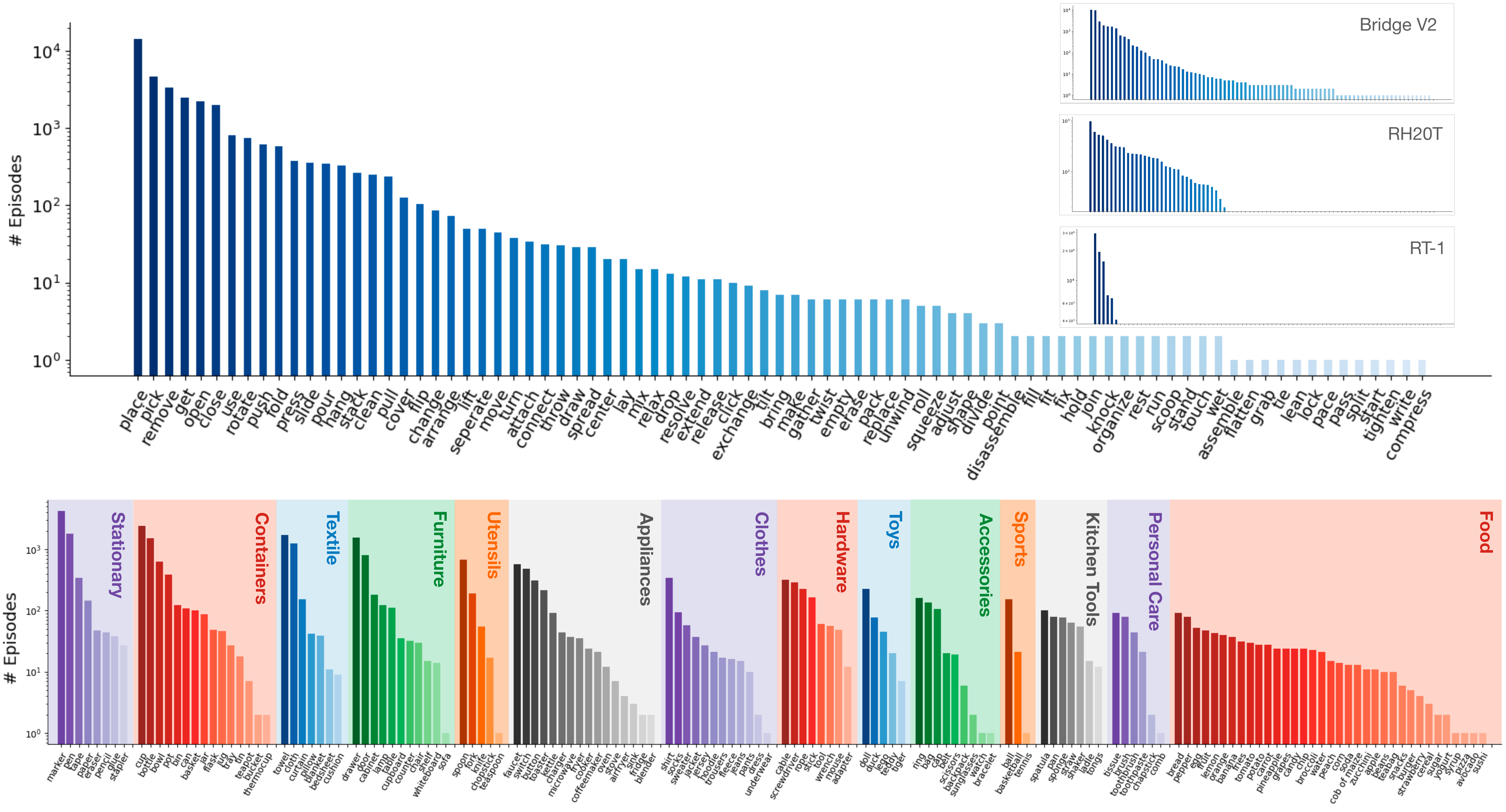}
  \caption{\small Distribution of verbs and objects in \name{}. \textbf{Top}: Distribution of verbs after de-duplication with GPT-4. \name{} has a long tail of diverse tasks that span a wide range of behaviors. We also visualize the verb distributions for existing large manipulation datasets and find that only Bridge~V2~\citep{walke2023bridgedata} has a comparable long tail of skills (for a detailed view of verb distributions for all datasets, see Appendix, \cref{fig:verb_distribution}). \textbf{Bottom}: Distribution of objects the robot interacts with in \name{}, sorted by category (best viewed zoomed in; for a detailed view, see \cref{fig:object_distribution}).
  }
  \label{fig:verb_object_distribution}
\end{figure*}

A crucial component of building the \name{} dataset was distributed data collection at \ninstitutions{}~institutions around the world: it is what enabled us to collect manipulation data across a large diversity of scenes and tasks. A key challenge in this distributed setup is robot hardware: how can we ensure \emph{consistent and reproducible} robot control across so many setups, locations and time zones? To streamline the distributed data collection process we designed the \name{} robot platform (see \cref{fig:robot_setup}), a hardware platform for data collection that is shared between all institutions, allowing us to quickly set up new data collection units and roll out updates across the whole data collection fleet. It is designed to support easy transportation between scenes and quick adjustment to new scenes and tasks.

We chose the Franka Emika Panda 7~DoF robot arm as the base of our setup since it is widely adopted in the robot research community, reliable, relatively affordable and was available at most participating institutions.
The robot arm is equipped with a Robotiq 2F-85 gripper and is mounted on a height-adjustable standing desk with wheels so it can easily move between scenes and buildings. We record image observations with three synchronized stereo camera streams: two exterior Zed 2 cameras, table-mounted on adjustable tripods to quickly adapt to a new scene layout, and a wrist-mounted Zed-Mini camera. We use the Polymetis controller~\citep{Polymetis2021} and record actions both in robot joint space and in end-effector space at a control frequency of 15Hz. 
The setup is completed with the Franka robot control box, a NUC that hosts the Polymetis server and an Alienware laptop that runs our data collection GUI (see \cref{sec:data_collect_protocol}). Everything is powered with a single power cable to further simplify changes in location.

For teleoperation, we use the controllers of a Meta Quest 2 headset to control the pose of the arm in 6D space as well as the gripper in continuous space. Over the course of this project we have replicated this setup \nrobots{}~times across various locations in North America, Asia, and Europe. We provide a thoroughly tested guide to replicate the hardware and software of our setup. We found that the setup is well-suited for data collection and policy learning across a wide range of scenes and tasks.

\subsection{Data Collection Protocol}
\label{sec:data_collect_protocol}

Our dataset is collected by \ncollectors{}~data collectors across various research institutions. A shared data collection protocol helps streamline data collection, particularly for inexperienced data collectors. When designing the collection protocol for \name{}, we focused on the following objectives: (1)~preventing common data collection mistakes like ``camera cannot see robot'' or ``teleoperator in camera view'', (2)~encouraging collection of diverse data, (3)~allowing data collectors to creatively choose scenes and tasks.

Every data collection session starts with moving the robot to a new scene. Data collectors were encouraged to choose scenes that include multiple interesting tasks, numerous interaction objects, and a healthy amount of clutter (see example scenes in \cref{fig:quali_data_examples}).
After setting up the robot in the new scene, the data collector chooses views for the 3rd person cameras that can capture a wide range of interesting behaviors in the scene. Then they perform extrinsic camera calibration using a checkerboard and the OpenCV calibration algorithm. Next, the data collector will enter all potential tasks for the current scene into a data collection GUI on the laptop attached to the robot, either by selecting from a list of task options or by typing in free-from task instructions (see \cref{fig:gui} for screenshots of the GUI). During data collection the GUI will prompt the data collector with a \emph{randomly} sampled task from this list for each new episode. This way we ensure that there is high coverage of diverse tasks and collection is not biased to easier tasks or closer objects. 
Additionally, the GUI periodically prompts the data collector to perform randomly sampled ``scene augmentations'' like nudges to the mobile base, moving and re-calibrating the 3rd person cameras, changing the room lighting, and adding or removing items within the scene. For each trajectory, we record the output of all RGB cameras, relevant low level state information from the robot, equivalent robot control commands from various popular action spaces, a data collector ID, and the metadata entered in the GUI~(see \cref{sec:collected_data_features} for a detailed list of all features we record). The data collector also marks whether the collected sequence was a success, which we log as part of the metadata. \name{} consists of \ntrajs{}~successful episodes; roughly \nfailtrajs{}~trajectories in our data collection were labeled as ``not successful'', which we include in our dataset release but \emph{do not} count towards the size of \name{}. A data collector will typically collect up to 100~trajectories or about 20~minutes of interaction data per scene before moving on to a new scene.

During post-processing, we label each episode with natural language commands using crowdsourcing via the \href{www.tasq.ai}{\url{tasq.ai}} data labeling platform. We provide up to three independently labeled instructions per episode from different crowd workers to ensure diversity of annotations.

Since the initial extrinsic calibration parameters, provided through conventional calibration detailed above, may not always be accurate due to factors such as checkerboard misalignment, inconsistent lighting, or errors inherent to the OpenCV calibration method, we address these inaccuracies in \cref{subsec:auto_camera_calib}. We discuss in detail the automatic post-hoc calibration process and provide three comprehensive sets of camera calibration matrices for the DROID dataset, each accompanied by respective quality assessment metrics. These include camera-to-base calibrations for around 36k unique scenes with one camera calibrated relative to the base, camera-to-camera calibrations for all scenes, and a curated superset of 24k scenes covering all three calibration methods with both cameras calibrated relative to the base. These refined calibrations enhance the dataset's suitability for robust geometric understanding in robotics and 3D perception tasks. For more details, please see Sec.~\cref{subsec:auto_camera_calib}.

\section{\name{} Dataset Analysis}
\label{sec:data_analysis}

While we have so far referred to \name{} and other large-scale robot manipulation datasets as ``diverse,'' there is nuance in what constitutes a diverse robot dataset. 
Different axes of data diversity will affect the generalization abilities of models trained on the data differently: scene diversity may facilitate generalization to new scenes, while task or camera viewpoint diversity allows for greater generalization to new instructions and camera angles. We will analyze \name{} along multiple important axes of diversity and compare it to existing large robot manipulation datasets.

When deciding which axes of generalization to inspect for robot manipulation datasets, it is important to consider which aspects of the problem may change between the training and downstream usage scenarios, \ie which axes we want manipulation policies to generalize over. This may involve aspects of the scene, task, and robot setup. We identify the following important axes of diversity for closer analysis: task diversity, object diversity, scene diversity, viewpoint diversity, and interaction location diversity. The latter refers to the diversity of 3D locations relative to the robot's base at which interactions with objects occur, an important factor when generalizing to new scene layouts where interactions often need to generalize to new table heights or new parts of the robot's workspace. 

We analyze \name{} along these axes and compare it to existing large-scale robot manipulation datasets~\citep{walke2023bridgedata,brohan2022rt,fang2023rh20t}. %
For each dataset, we run our analysis using one randomly sampled third-person camera frame per episode %
and the provided language instruction annotations. We find that results are consistent across randomly sampled frames.

\begin{figure}[t]
  \centering
  \includegraphics[width=1.0\linewidth]{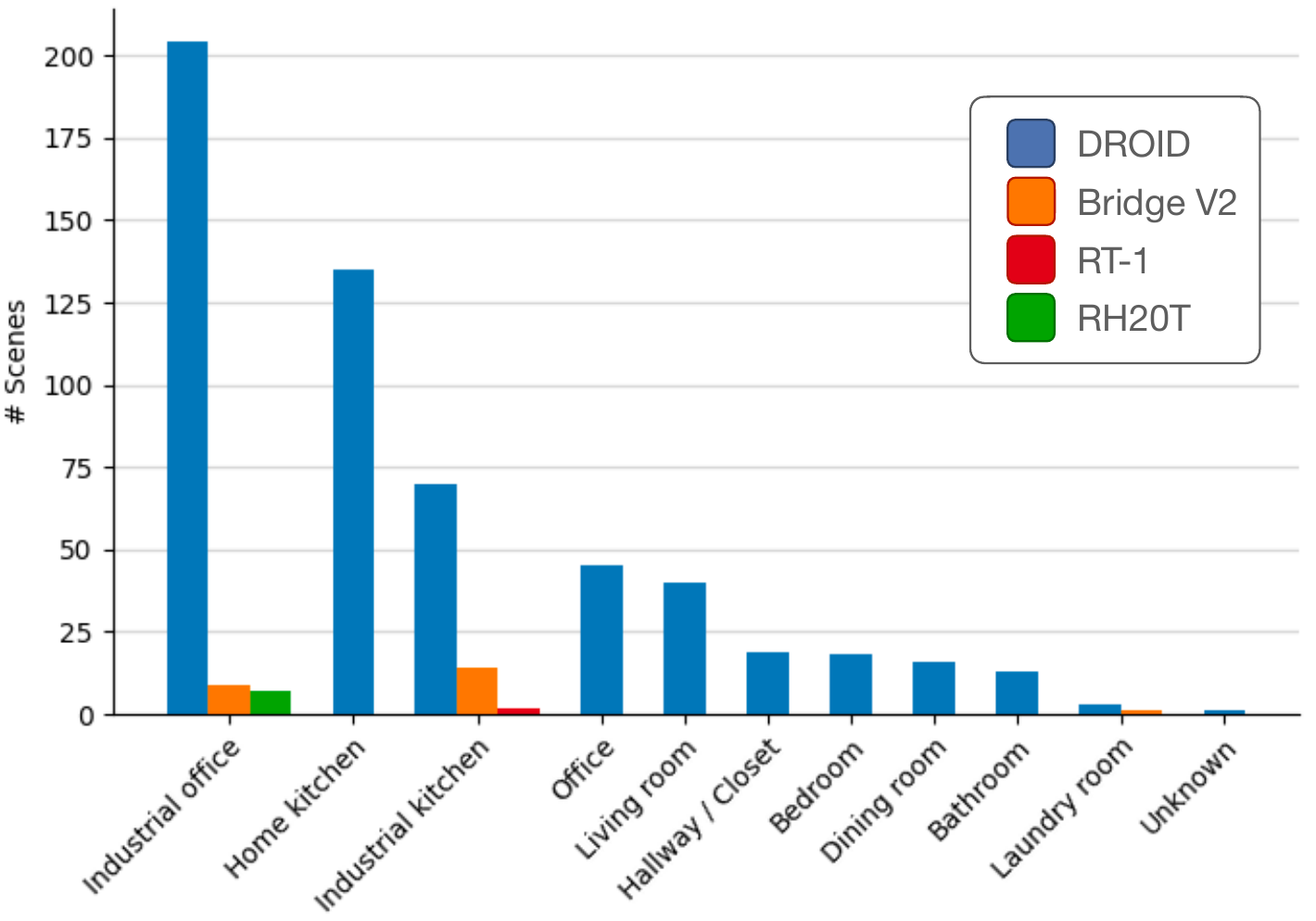}
  \caption{
    \small Number of scenes per scene type. \name{} has an order of magnitude more scenes than other large robot manipulation datasets, spanning a much wider range of scene types. We manually verified or confirmed with the authors that scene count and type for prior datasets is accurately reported.
  }
  \label{fig:scene_distribution}
\end{figure}

\begin{figure}[t]
  \centering
  \includegraphics[width=1.0\linewidth]{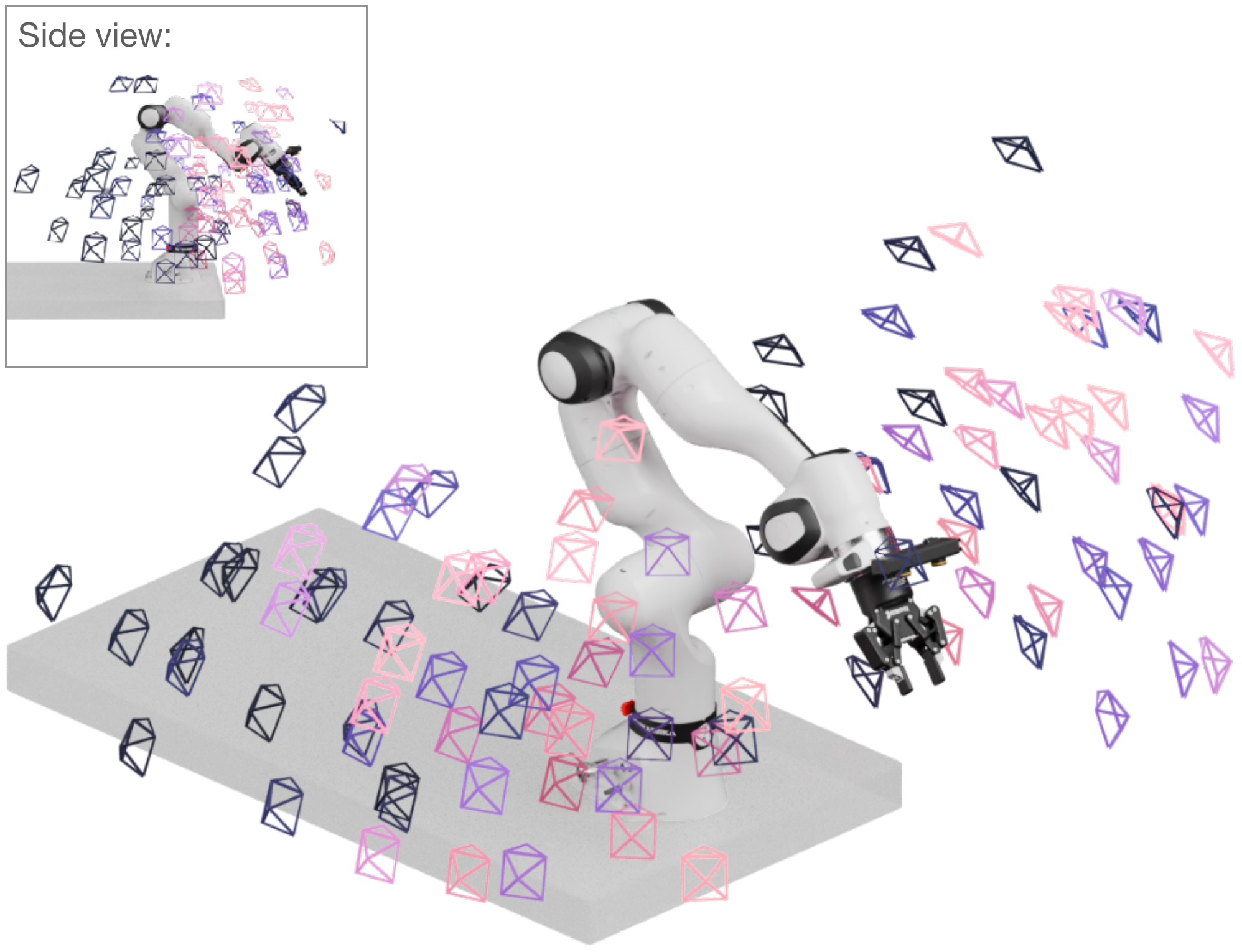}
  \caption{
    \small Third-person camera viewpoints in \name{} (subsampled). \name{} episodes cover a total of \ncamposes{}~camera viewpoints along with intrinsic and extrinsic stereo camera calibration. Brighter colors indicate regions of higher viewpoint density.
  }
  \label{fig:viewpoint_distribution}
\end{figure}

\begin{figure}[t]
  \centering
  \includegraphics[width=1.0\linewidth]{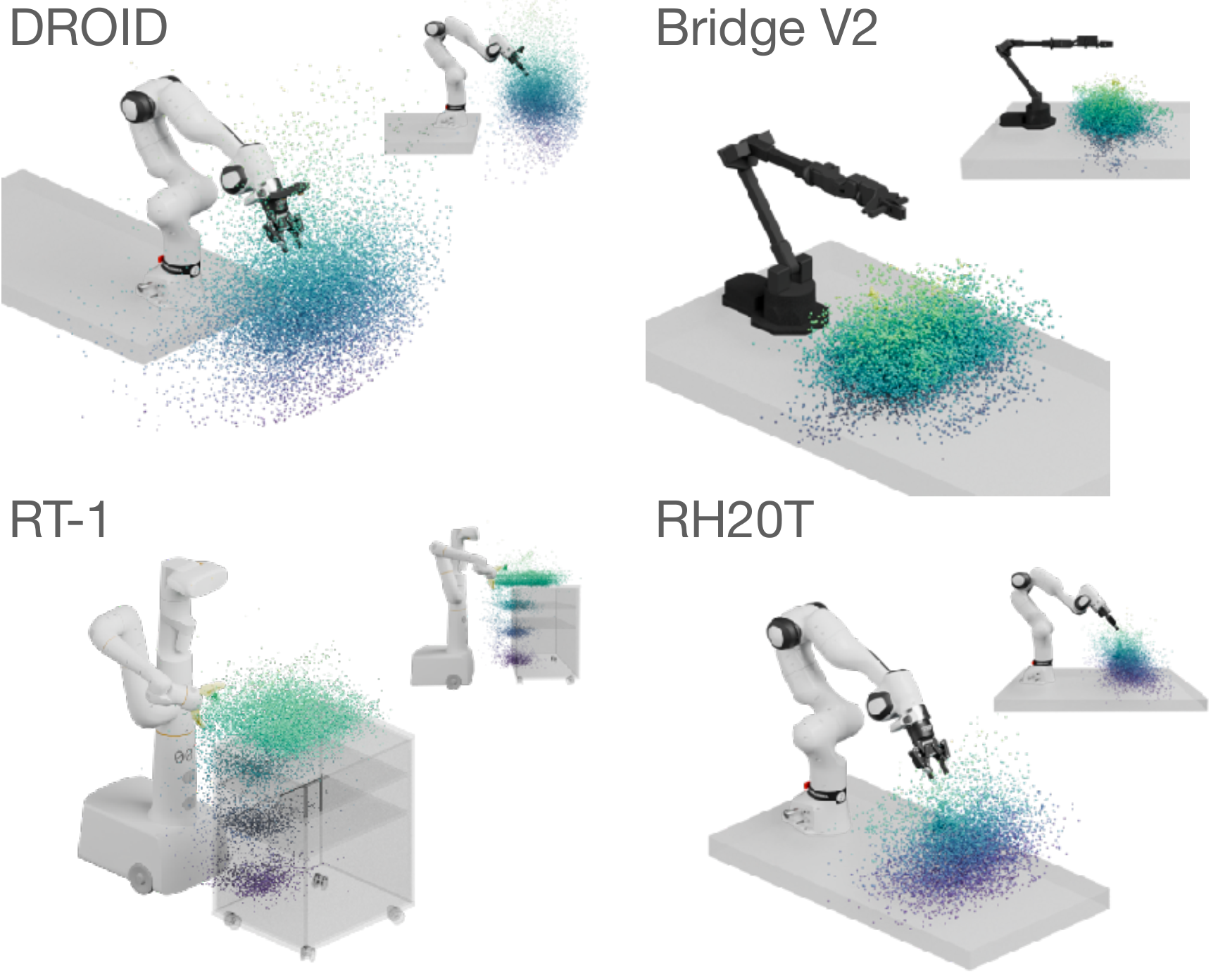}
  \caption{\small Visualization of 3D interaction points relative to the robot base. We visualize the 3D location at which the gripper first closes in each trajectory, since closing the gripper often indicates meaningful object interactions. \name{}'s interactions cover a larger part of the robot's workspace, since the robot is moved freely between collection sessions instead of being placed in front of repetitive tabletop scenes.}
  \label{fig:interact_point_viz}
\end{figure}

We visualize the results of our analysis in \cref{fig:verb_object_distribution,fig:scene_distribution,fig:viewpoint_distribution,fig:interact_point_viz}. \textbf{Overall, we find that \name{} significantly increases diversity in tasks, objects, scenes, viewpoints and interaction locations over existing large scale robot manipulation datasets}. 
A key reason is \name{}'s data collection protocol (see \cref{sec:data_collect_protocol}): by collecting data with \ncollectors{}~data collectors in \nbuildings{}~buildings across three continents, switching scenes approximately every 20~minutes during collection and giving collectors the freedom to freely choose scene-appropriate tasks,
we can substantially increase the diversity of scenes, tasks, and objects featured in the dataset. Next, we will describe our analysis for each category in more detail.

\paragraph{Task diversity} As explained in \cref{sec:related_work}, we use the distribution of de-duplicated verbs in a dataset's instructions as a scalable indicator for behavioral diversity. We use a semantic parsing algorithm~\citep{spacy2} to extract verbs and referenced objects from the language instructions. We then use GPT4 to de-duplicate the verbs, \ie remove synonyms and typos. We plot the distribution of verbs for \name{} in \cref{fig:verb_object_distribution}, top. 
\name{} features a wide variety of verbs with a long-tailed distribution. We use a logarithmic scale for these visualizations, since diversity is about covering a wide range of tasks, rather than having a high concentration of many episodes on only a handful of tasks -- that is, it is less important if a task has 1000 vs. 2000 trajectories than whether it has 0 vs. 10. We also visualize the corresponding verb distributions for existing large manipulation datasets~\citep{walke2023bridgedata,fang2023rh20t,brohan2022rt}, and find that only Bridge~V2~\citep{walke2023bridgedata} has a comparable long tail of verb classes, although in a more restricted set of scenes (see scene diversity analysis below). \cref{fig:verb_distribution} shows a detailed view of the verb distributions for all datasets.

\paragraph{Object diversity}
A dataset that includes manipulations for a large variety of objects facilitates generalization to new objects downstream. We analyze the objects the robot manipulates for each episode in \name{} from the language instruction labels using the same semantic parsing pipeline~\citep{spacy2} and show the distribution in \cref{fig:verb_object_distribution}, bottom (best viewed zoomed in, or see \cref{fig:object_distribution} for an enlarged version). \name{} contains interactions with a wide range of everyday objects, spanning a diverse set of categories. %
We also plot the \emph{joint} distribution of the most common verbs and interacted objects in \cref{fig:verb_object_heatmap}. It shows that \name{} not only contains diverse objects, but also a diverse range of interactions with most objects.

\paragraph{Scene diversity} We define 10~scene types (see \cref{fig:scene_distribution}) and use GPT-4V to determine the scene type for a given episode in \name{} (see Appendix~\ref{sec:scene_type_classification} for the used prompt). We find that this leads to high-quality scene type annotations (see \cref{fig:quali_data_examples} for example scenes and their categorization). For existing robot datasets we manually determine the scene types for each scene due to the small number of total scenes. 
\name{} contains \nscenes{}~unique scenes, an order of magnitude more than existing large robot manipulation datasets. The scenes cover a wide spectrum of scene types, from office environments to households. Qualitatively, the scenes in \name{} reflect realistic real world scenarios with naturally occuring objects and backgrounds. We highly encourage the reader to inspect qualitative examples of scenes in \cref{fig:quali_data_examples} and the supplementary videos.

\paragraph{Viewpoint diversity} Existing large-scale robot learning datasets often only contain a limited set of camera viewpoints because the cameras are mounted in a fixed location relative to the scene or robot. %
In contrast, \name{} varies camera viewpoints significantly during data collection and thus has a broad coverage of viewpoints (see \cref{fig:viewpoint_distribution}) with \ncamposes{}~unique view points in the dataset. %

\begin{figure*}[t]
  \centering
  \includegraphics[width=1.0\linewidth]{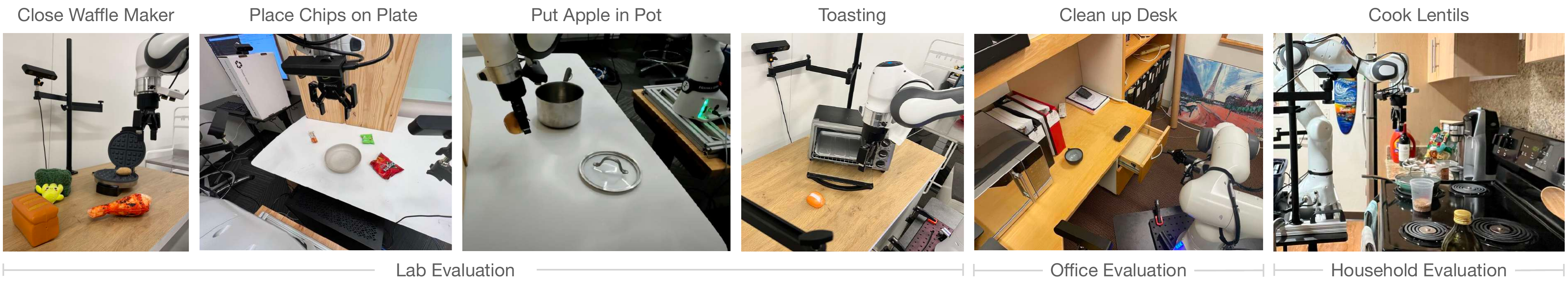}
  \caption{\small Robot setups for policy evaluation. We cover a wide range of tasks and scenes, from lab evaluations to offices and real households, to reflect the diversity of use cases in real robot research. Depending on the task we collect between 50 and 150 demonstrations. We describe each task with out-of-distribution evaluation modifications in parenthesis, left to right: \textbf{Close Waffle Maker}: The robot needs to close a waffle maker (distractor objects). \textbf{Place Chips on Plate}: The robot needs to pick up the chips bag and place it on the provided plate (unseen chips bag and distractor objects). \textbf{Put Apple in Pot}: The robot needs to pick up the apple, place it in the pot, and close the lid (unseen distractor object). \textbf{Toasting}: The robot needs to pick up the object, place it in the toaster oven, and close the oven (toast a novel object). \textbf{Clean up Desk}: The robot needs to open the drawer, place the eraser that is on top of the desk inside the drawer, and close it (distractor objects on desk and in drawer). \textbf{Cook Lentils}: The robot needs to remove the pan lid, pick up and pour lentils into the pan, and turn on the stove(add distractor objects).
  }
  \label{fig:eval_tasks}
\end{figure*}
\paragraph{Interaction location diversity} Another subtle yet important aspect of robot datasets is the diversity of interaction locations: are tasks always executed in the same narrow slice of the workspace, \eg at the same table height, or does the data cover interactions across a large fraction of the work space? We use the point of first gripper closing in every episode as a proxy for interactions in the dataset and visualize the 3D location of these interaction points for different datasets in \cref{fig:interact_point_viz}. \name{} features interactions in a wider range of the workspace than existing robot manipulation datasets that typically focus interactions on a table surface in front of the robot. %

\section{Experiments}
\label{sec:experiments}

The analysis in the previous section highlighted the diversity of tasks, objects, scenes, and viewpoints in the \name{} dataset. 
In this section, we investigate whether this diverse data resource can be used to boost policy performance and robustness across a wide spectrum of robot manipulation tasks and environments. To this end, we train policies across \nevaltasks{}~tasks in \nevallocations{}~different locations including lab, office, and household settings, to reflect the diversity of real world robotic research use cases (see \cref{fig:eval_tasks}). All experiments use representative, state of the art robot policy learning approaches~\citep{chi2023diffusionpolicy}. Across the board, we find that \name{} improves policy success rate while increasing robustness to scene changes like distractors or novel object instances.

\begin{figure*}[t]
  \centering
  \includegraphics[width=1.0\linewidth]{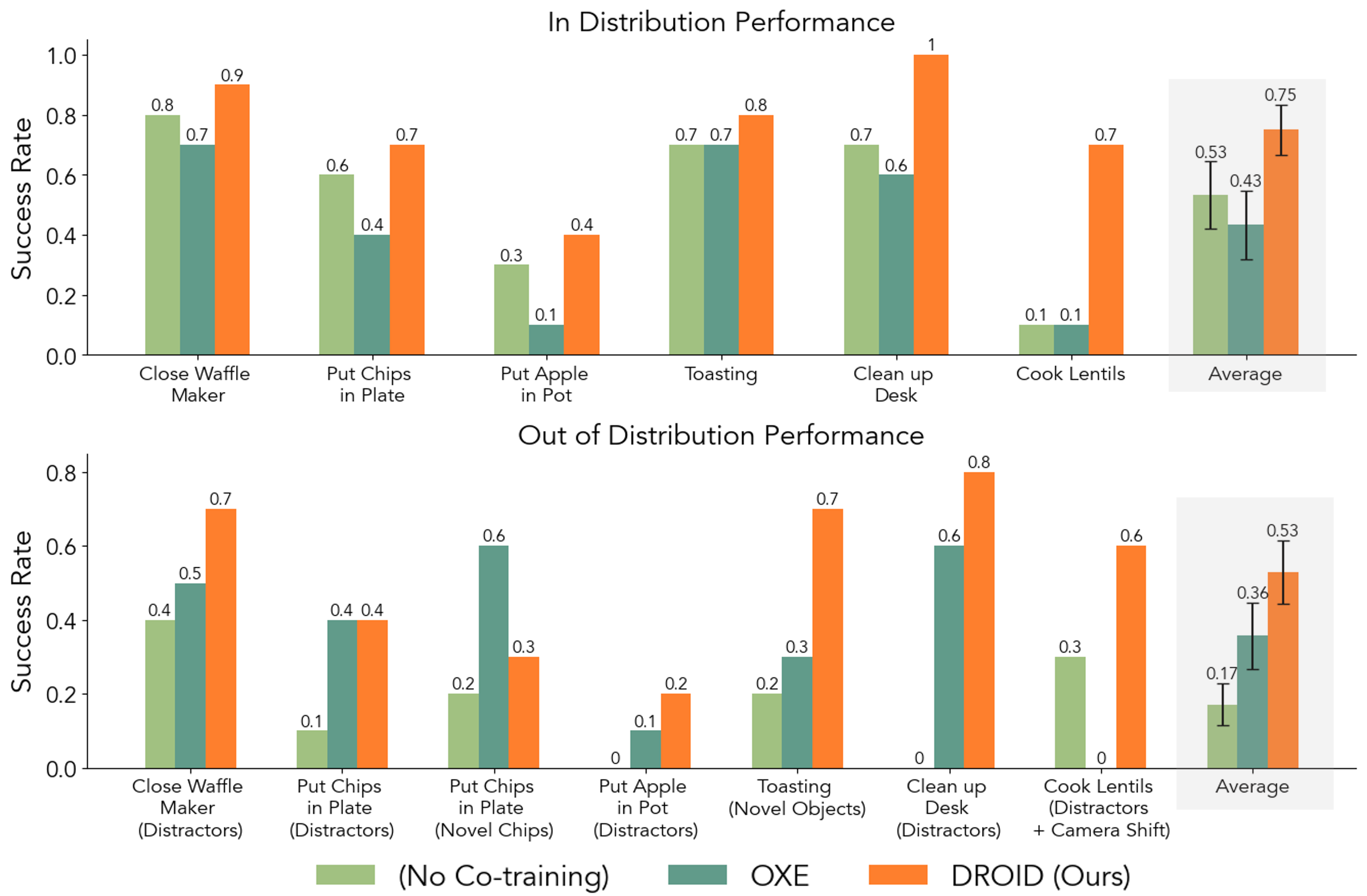}
  \caption{\small \textbf{Does \name{} Improve Policy Performance and Robustness?} We find that across all our evaluation tasks, co-training with \name{} significantly improves both in distribution and OOD performance over both no co-training and co-training with the Open-X dataset. We compare success rate averaged across all tasks with standard error, and find \name{} outperforms the next best method \textbf{by 22\%} absolute success rate in-distribution and \textbf{by 17\%} out of distribution.
  }
  \label{fig:cotrain_fig}
\end{figure*}
\subsection{Experimental Setup}
\label{sec:exp_setup}

\paragraph{Tasks}
As illustrated in \cref{fig:eval_tasks}, we choose \nevaltasks{}~tasks in \nevallocations{}~locations that span a representative range of real robot learning use cases: from simple pick-place tasks to multi-stage cooking tasks; from clean lab settings to real households. All experiments use the \name{} hardware stack for policy evaluations. Concretely, we evaluate on the following \nevaltasks{}~tasks, each with their own out-of-distribution variants:

\noindent \textbf{Closing Waffle Maker}: A short horizon task in a lab setting (70 demonstrations), where the task is to close a waffle maker. The waffle maker position is randomized between episodes. The out of distribution variant consists of adding several distractor objects on the table.  

\noindent \textbf{Place Chips on Plate}: A short horizon task in a lab setting (50 demonstrations), where the task is to pick and place a bag of Doritos chips onto a plate, with two distractor objects on the table. All objects and the plate position are randomized between episodes on the table. The out of distribution variant consists of (a) changing the type of chips or (b) adding more distractor objects to the table. 

\noindent \textbf{Put Apple in Pot}: A medium horizon task in a lab setting (60 demonstrations), where the task is to pick and place an apple into a pot and then put a lid on the pot. The apple, pot, and lid position are randomized between episodes on the table. The out of distribution variant involves placing a distractor plate on the table. 

\noindent \textbf{Toasting}: A medium horizon task in a lab setting (150 demonstrations), where the task is to put an object on a toaster oven tray, then close the toaster oven. The object and toaster position are randomized between episodes on the table. The out of distribution variant consists of toasting novel objects. 

\noindent \textbf{Clean up Desk}: A long horizon task in an office setting (50 demonstrations), where the task is to open a drawer, pick and place an eraser into the drawer, and then close the drawer. The eraser position is fixed. The out of distribution variant consists of adding distractor objects on the desk and in the drawer. 

\noindent \textbf{Cook Lentils}: A long horizon task in a kitchen setting (50 demonstrations), where the task is to remove the lid off a pan, pour lentils into the pan, and turn on the stove. The object positions are fixed. The out of distribution variant consists of adding several distractor objects and a camera shift.

Additional details about each evaluation task can be found in Appendix~\ref{sec:eval_procedure}.  All data is collected using the \name{} teleoperation setup and training uses the same standardized policy learning backbone.

\begin{figure*}[t]
  \centering
  \includegraphics[width=1.0\linewidth]{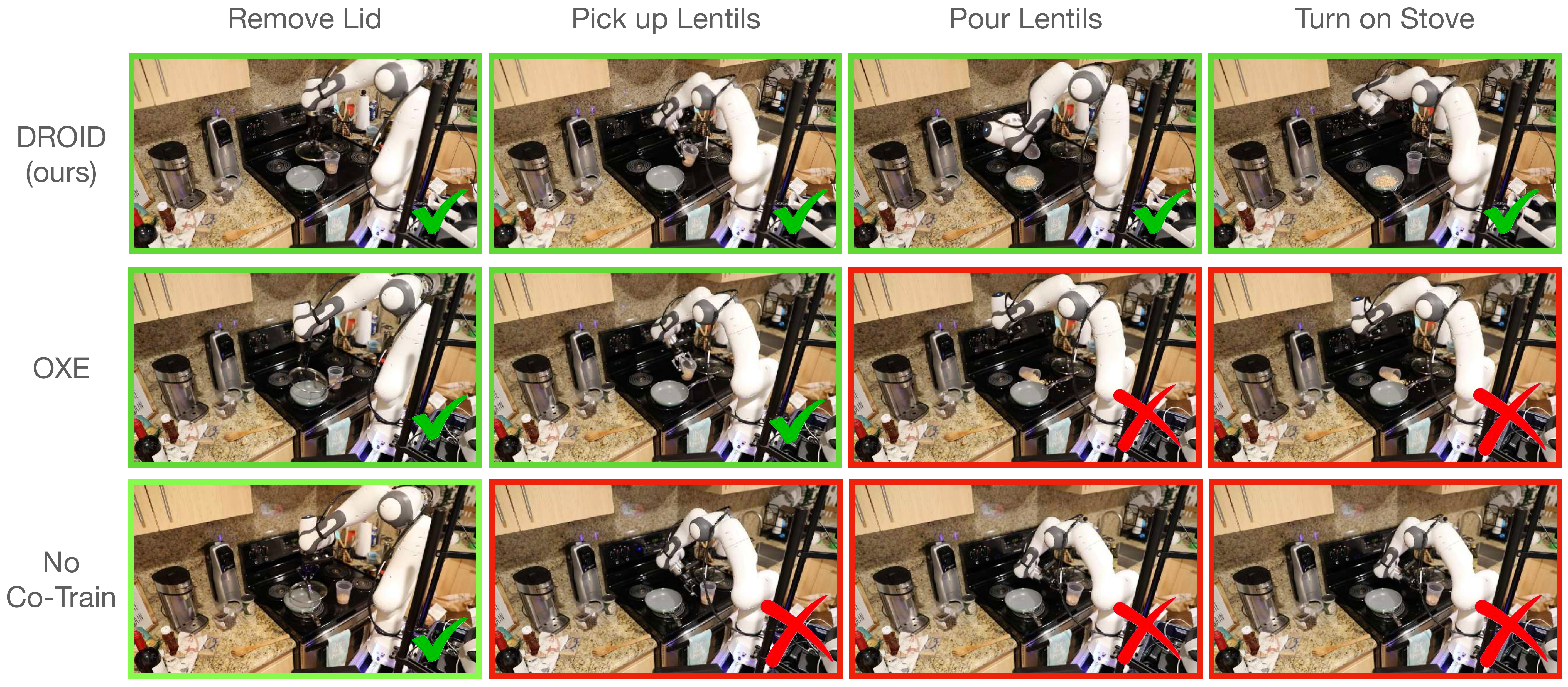}
  \caption{\small Representative policy rollout examples on the most challenging ``Cook Lentils'' task in the kitchen of one of the authors. Qualitatively, we find that policies co-trained with \name{} perform smoother, more precise motions, allowing them to solve long-horizon tasks like lentil cooking even in the presence of unseen distractor objects. In contrast, policies trained only with in-domain demonstrations or co-trained with Open-X data~\citep{open_x_embodiment_rt_x_2023} struggle with long-horizon tasks and out-of-distribution evaluation settings. %
  See \website~for rollout videos.
  }
  \label{fig:quali_rollout}
\end{figure*}

\paragraph{Policy training}
The goal of this work is to introduce a new robot manipulation dataset, \emph{not} to introduce a new policy learning method. Thus, during experimental evaluations we aim to leverage a well-adopted, state-of-the-art policy learning pipeline. To this end, we use diffusion policies~\citep{chi2023diffusionpolicy,ha2023scaling,sridhar2023nomad,octo_2023,fu2024mobile}, which leverage denoising diffusion models for action prediction and have recently demonstrated strong performance across a range of applications. We build on the implementation of diffusion policies in Robomimic~\citep{robomimic2021}, which provides high quality open-source implementations of a number of different imitation learning and offline RL algorithms. Concretely, all of our policies are conditioned on a language instruction, use the RGB camera streams from the two external cameras and the robot proprioception as input, and produce absolute robot end-effector translation, rotation, and gripper actions. We first downsample the camera observations to a resolution of $128 \times 128$ and use a ResNet-50 visual encoder pre-trained on ImageNet~\citep{deng2009imagenet} to encode both visual inputs. We then concatenate these visual embeddings with a frozen DistilBERT~\citep{Sanh2019DistilBERTAD} language embedding and the robots proprioceptive state. These concatenated features are then fed through an MLP and passed to a U-Net diffusion head which generates action trajectories. In line with prior work~\citep{chi2023diffusionpolicy}, we train the diffusion policy to generate 16-step action sequences, and during rollouts, step 8 actions open loop before re-running policy inference. For leveraging \name{} during policy training, we simply mix training batches at a 50/50 ratio between the small in-domain dataset and the complete \name{} dataset but excluding trajectories marked as ``not successful'', which we find to work well in practice. 
Additional details about the policy training can be found in Appendix~\ref{sec:policy_details}.

\subsection{Does \name{} Improve Policy Performance and Robustness?}
\label{exp:co_training}

To study if co-training with \name{} can enable improved policy learning, we 
train separate policies for each evaluation task and compare all policies head-to-head in A/B evaluations using 10~rollouts for each task setting and method. To test how \name{} and existing datasets affect policy robustness, we evaluate each task and method in two settings: ``in-distribution,'' which reflects the distribution of tasks in the in-domain demonstrations with noise added to the initial robot and object positions, and ``out-of-distribution'' (OOD), which tests policy robustness \eg by introducing distractor objects or switching the manipulated object. We evaluate the following approaches:
\begin{itemize}
    \item \textbf{No Co-training}: Trains a diffusion policy~\citep{chi2023diffusionpolicy} using the in-domain demonstrations only
    \item \textbf{\name{} (Ours)}: Trains a diffusion policy, but mixes batches 50/50 between in-domain demonstrations and \name{} demonstrations
    \item \textbf{OXE~\citep{open_x_embodiment_rt_x_2023}}: Trains a diffusion policy, but mixes batches 50/50 between in-domain demonstrations and trajectories from the Open X-Embodiment dataset~\citep{open_x_embodiment_rt_x_2023} (OXE). %
    OXE contains most of the existing large robot manipulation datasets we compared \name{} to in \cref{sec:data_analysis}, as well as a large number of other robot datasets, spanning 22~robot embodiments and approximately 300~scenes total.\footnote{We use a curated split of OXE based on \citet{octo_2023}, which has been shown to work well for policy learning in prior work~\citep{octo_2023}. We remove the Language Table dataset~\citep{lynch2023interactive}, equivalent to 5\% of the Octo training mix, due to its repetitive scene layouts and tasks, and its raw size, which proved challenging to handle for our training infrastructure.}
\end{itemize}

We present the results of our policy evaluations in \cref{fig:cotrain_fig}. Across all tasks, we find that \name{} substantially improves policy performance compared to the diffusion policy trained on in-domain data only. Policies co-trained with \name{} also perform better than policies that leverage diverse, existing robot datasets in Open X-Embodiment (OXE). Notably, when testing out of distribution performance, the No Co-training baseline performs quite poorly while the co-trained policies are much more effective. This difference is especially notable when co-training with \name{}, which has the strongest overall performance.

Qualitatively, we find that policies that leverage \name{} during training are notably smoother and precise than other comparisons, particularly in the more challenging out-of-distribution evaluations. For instance, in the OOD setting of the Waffle Closing task, \name{} is the only method that consistently reaches for the waffle maker, while the other methods get confused about the task. Similarly, in the multi-step Cook Lentils task, baselines tend to fail after two or sometimes just one step, while co-training with \name{} is the only method able to consistently finish all three steps See \cref{fig:quali_rollout} for examples of qualitative task rollouts.

\subsection{How important is the scene diversity in \name{}?}
\label{exp:scene_div}

\begin{figure}[t]
  \centering
  \includegraphics[width=1.0\linewidth]{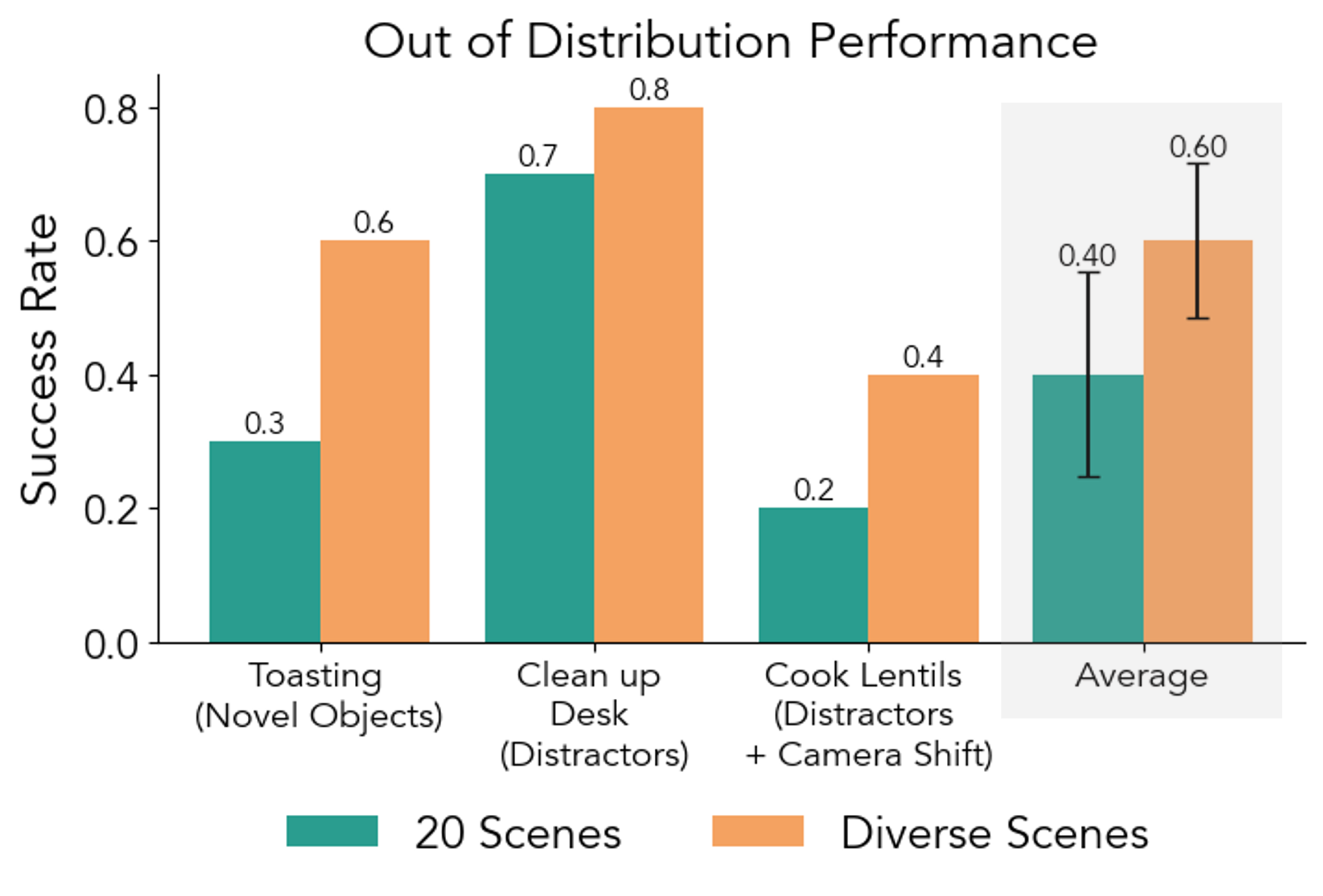}
  \caption{\small \textbf{How important is the scene diversity in \name{}?} We find that co-training on a subset of \name{} with diverse scenes has higher OOD performance than co-training on a subset of \name{} with only 20 scenes, suggesting the scene diversity of \name{} is one of the driving factors behind strong policy performance when co-training. 
  }
  \label{fig:filter_fig}
\end{figure}

One of the unique benefits of \name{} compared to existing robot datasets is its amount of \emph{scene diversity}. Indeed we see in Figure~\ref{fig:scene_distribution} that \name{} contains far more scene diversity than the next most diverse robot manipulation dataset. While we've seen the benefits of co-training with \name{}, \emph{can we quantify how much of a role scene diversity plays in improved policy robustness?}

To test this, we design an experiment that uses the challenging OOD versions of the evaluation tasks from Section~\ref{sec:exp_setup}, but compares:

\begin{itemize}
    \item \textbf{\name{} (7k, 20 Scenes)}: Selects for the 20 scenes from \name{} with the most demonstrations each, resulting in 7362 trajectories with comparatively little scene diversity. 
    \item \textbf{\name{} (7k, Diverse Scenes)}: Uniform random sample of 7362 successful demonstrations from the \name{} dataset, which matches dataset size to the previous method while retaining high scene diversity. 
\end{itemize}

These comparisons use the same 50/50 co-training paradigm with individual task data used in the previous experiment. Hence, this helps establish whether the scene diversity of \name{} results in better policy performance than just using 20 scenes while controlling for dataset size.

In Figure~\ref{fig:filter_fig} we observe that using the split of the dataset with more diverse scenes yields better performance in the OOD evaluation setting. By comparing Figure~\ref{fig:filter_fig}'s individual task performances with the corresponding tasks in Figure~\ref{fig:cotrain_fig}, we also see that the performance of co-training with the full \name{} dataset matches or outperforms the performance with the subsampled dataset on all three tasks.
These results suggest that the strength of \name{} lies in its size and especially in its \emph{diversity}.

\section{Discussion}
\label{sec:conclusion}

In this work, we introduced \namelong{}, a new robot manipulation dataset with a large diversity of scenes, tasks, objects and viewpoints. Our dataset analysis in \cref{sec:data_analysis} showed that \name{} has an order of magnitude larger scene diversity than existing large robot manipulation datasets, a wide range of tasks, many interaction objects, and diverse viewpoints. Our policy learning evaluations show that \name{} is a valuable data resource for improving policy performance and robustness, even in comparison to existing large robot data sources like the Open X-Embodiment dataset~\citep{open_x_embodiment_rt_x_2023}.

We hope that \name{} can be a catalyst for research on general-purpose robot manipulation policies that are able to generalize to a broad range of tasks and scenes. In this work, we showed one example for leveraging \name{} to boost policy performance, but there are many open questions about how to best make use of such diverse data: how should we combine \name{} with existing large-scale robot datasets and how can we train policies that perform tasks in new scenes without \emph{any} in-domain data? Can the diverse interaction data in \name{} be used to learn better visual representations for robotic control? And in what situations is it helpful to train on the full dataset vs. slices of the data? We hope that \name{} can accelerate research on these questions and are excited for how the community will leverage the dataset! We also hope that our open-sourced hardware platform, which already exists in \nlabs{}~labs around the globe and is easy to reproduce, can improve reproducibility of robot learning research and facilitate future additions to the \name{} dataset.

\section*{ACKNOWLEDGMENT}

We thank the Toyota Research Institute (TRI) for their support in various aspects of this project, from data collection to compute for policy training. This work was supported by the Google TPU Research Cloud. We further acknowledge the following funding sources: Chelsea Finn's group was supported by TRI and ONR grants N00014-20-1-2675 and N00014-22-1-2621; Sergey Levine's group was supported by TRI, NSF FRR IIS-2150826, and ONR N00014-20-1-2383; Ram Ramamoorthy's group was supported by the United Kingdom Research and Innovation through grant EP/S023208/1 to the EPSRC Centre for Doctoral Training in Robotics and Autonomous Systems (RAS) and grant EP/V026607/1 to the UKRI Research Node on Trustworthy Autonomous Systems Governance and Regulation; Dorsa Sadigh's group was supported by TRI and ONR grant N00014-22-1-2293; Glen Berseth's group acknowledges funding support from NSERC and CIFAR and compute support from Digital Research Alliance of Canada, Mila IDT and NVidia; Jeannette Bohg's group was supported by TRI, Intrinsic, Toshiba and the National Science Foundation under Grant 2327974; Joseph Lim's group was supported by Institute of Information \& Communications Technology Planning \& Evaluation (IITP) grants (No.2019-0-00075, Artificial Intelligence Graduate School Program, KAIST; No.2022-0-00077, AI Technology Development for Commonsense Extraction, Reasoning, and Inference from Heterogeneous Data), and a National Research Foundation of Korea (NRF) grant (NRF-2021H1D3A2A03103683) funded by the Korean government (MSIT).

\bibliographystyle{plainnat}
\bibliography{bibref_definitions_long,bibtex}

\clearpage
\newpage

\begin{appendices}

\section{Contributions}
\noindent\textbf{Project Leads}: Alexander Khazatsky, Karl Pertsch\\[-0.1cm]

\noindent\textbf{Research Leads (contributed significantly to development of data collection setup, data post-processing and policy training)}: Alexander Khazatsky, Karl Pertsch, Suraj Nair, Ashwin Balakrishna, Sudeep Dasari, Siddharth Karamcheti, Soroush Nasiriany, Mohan Kumar Srirama\\[-0.1cm]

\noindent\textbf{Engineers (helped implement data collection, postprocessing and policy learning infrastructure)}: Alexander Khazatsky, Karl Pertsch, Suraj Nair, Ashwin Balakrishna, Siddharth Karamcheti, Soroush Nasiriany, Mohan Kumar Srirama, Lawrence Yunliang Chen, Kirsty Ellis, Peter David Fagan, Masha Itkina, Marion Lepert, Jason Ma, Patrick Tree Miller, Jimmy Wu, Huy Ha, Youngwoon Lee, Kaiyuan Wang, Kevin Black, Cheng Chi, Kyle Hatch, Shan Lin, Jingpei Lu, Abdul Rehman, Pannag R Sanketi, Cody Simpson, Quan Vuong, Blake Wulfe, Ted Xiao, Jonathan Yang, Arefeh Yavary, Tony Z. Zhao\\[-0.1cm]

\noindent\textbf{Lab Leads (coordinated data collection in their respective labs)}: Suraj Nair, Ashwin Balakrishna, Siddharth Karamcheti, Soroush Nasiriany, Mohan Kumar Srirama, Lawrence Yunliang Chen, Kirsty Ellis, Peter David Fagan, Joey Hejna, Masha Itkina, Marion Lepert, Jason Ma, Patrick Tree Miller, Jimmy Wu, Suneel Belkhale, Shivin Dass, Huy Ha, Abraham Lee, Youngwoon Lee, Arhan Jain, Marius Memmel, Sungjae Park, Ilija Radosavovic, Kaiyuan Wang, Albert Zhan, Archit Sharma, Homer Walke\\[-0.1cm]

\noindent\textbf{Policy Evaluators (ran robot evaluations for policy learning experiments)}: Alexander Khazatsky, Suraj Nair, Ashwin Balakrishna, Sudeep Dasari, Mohan Kumar Srirama, Joey Hejna, Donovon Jackson, Tony Nguyen, Derick Seale\\[-0.1cm]

\noindent\textbf{Data Collectors}: Alexander Khazatsky, Mohan Kumar Srirama, Lawrence Yunliang Chen, Kirsty Ellis, Peter David Fagan, Masha Itkina, Marion Lepert, Jason Ma, Patrick Tree Miller, Jimmy Wu, Suneel Belkhale, Shivin Dass, Abraham Lee, Arhan Jain, Marius Memmel, Sungjae Park, Ilija Radosavovic, Albert Zhan, Christopher Agia, Rohan Baijal, Mateo Guaman Castro, Daphne Chen, Qiuyu Chen, Trinity Chung, Jaimyn Drake, Ethan Paul Foster, Jensen Gao, David Antonio Herrera, Minho Heo, Kyle Hsu, Jiaheng Hu, Donovon Jackson, Charlotte Le, Yunshuang Li, Kevin Lin, Roy Lin, Zehan Ma, Abhiram Maddukuri, Suvir Mirchandani, Daniel Morton, Tony Nguyen, Abby O'Neill, Rosario Scalise, Derick Seale, Victor Son, Stephen Tian, Andrew Wang, Yilin Wu, Annie Xie, Jingyun Yang, Patrick Yin, Yunchu Zhang\\[-0.1cm]

\noindent\textbf{Lead Advisor}: Chelsea Finn\\[-0.1cm]

\noindent\textbf{Advisors}: Osbert Bastani, Glen Berseth, Jeannette Bohg, Ken Goldberg, Abhinav Gupta, Abhishek Gupta, Dinesh Jayaraman, Joseph J. Lim, Jitendra Malik, Roberto Martín-Martín, Subramanian Ramamoorthy, Dorsa Sadigh, Shuran Song, Jiajun Wu, Yuke Zhu, Thomas Kollar, Sergey Levine

\begin{figure*}[t]
  \centering
  \includegraphics[width=1.0\linewidth]{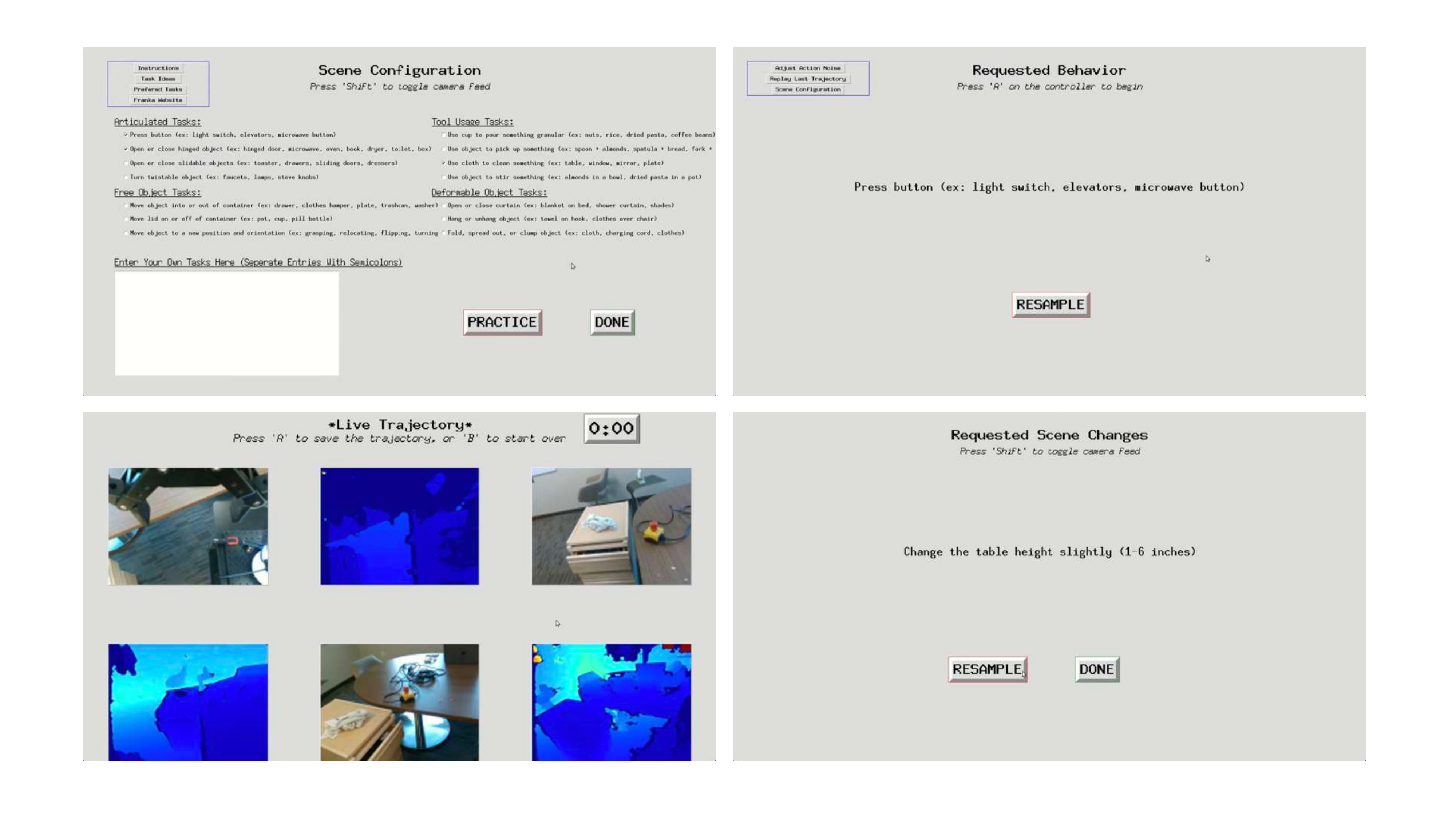}
  \caption{\name{} data collection GUI. \textbf{Top left}: Screen for entering feasible tasks for the current scene. Tasks can either be selected from a list of suggestions or typed as free-form instructions. \textbf{Top right}: Instruction screen -- the GUI samples a task \emph{at random} from the entered list of feasible tasks and instructs the data collector to record a demonstration trajectory. This ensures wide coverage of possible tasks in each scene and avoids bias towards easy or familiar tasks. \textbf{Bottom left}: Data collection screen -- displays RGB and depth camera live streams. \textbf{Bottom right}: The GUI periodically suggests scene changes between demonstration collections to ensure high scene diversity.
  }
  \label{fig:gui}
\end{figure*}

\section{\name{} Data Features}
\label{sec:collected_data_features}

All \name{} data is recorded at 15Hz. Each \name{} trajectory contains the following elements:
\begin{itemize}
    \item 3 stereo RGB camera streams at 1280x720 resolution
    \item robot joint positions and velocities (7D)
    \item robot end-effector pose and velocity in robot base frame (6D)
    \item robot gripper position and velocity (1D)
\end{itemize}
Additionally each trajectory has the following metadata:
\begin{itemize}
    \item 1-3 natural language instructions describing the task performed in the trajectory, collected via crowdsourcing
    \item extrinsic camera calibration matrices for both exterior cameras
    \item building name and data collector user ID
    \item scene type, as classified by GPT4V (see Section~\ref{sec:scene_type_classification})
\end{itemize}

\begin{figure*}[t]
  \centering
  \includegraphics[width=1.0\linewidth]{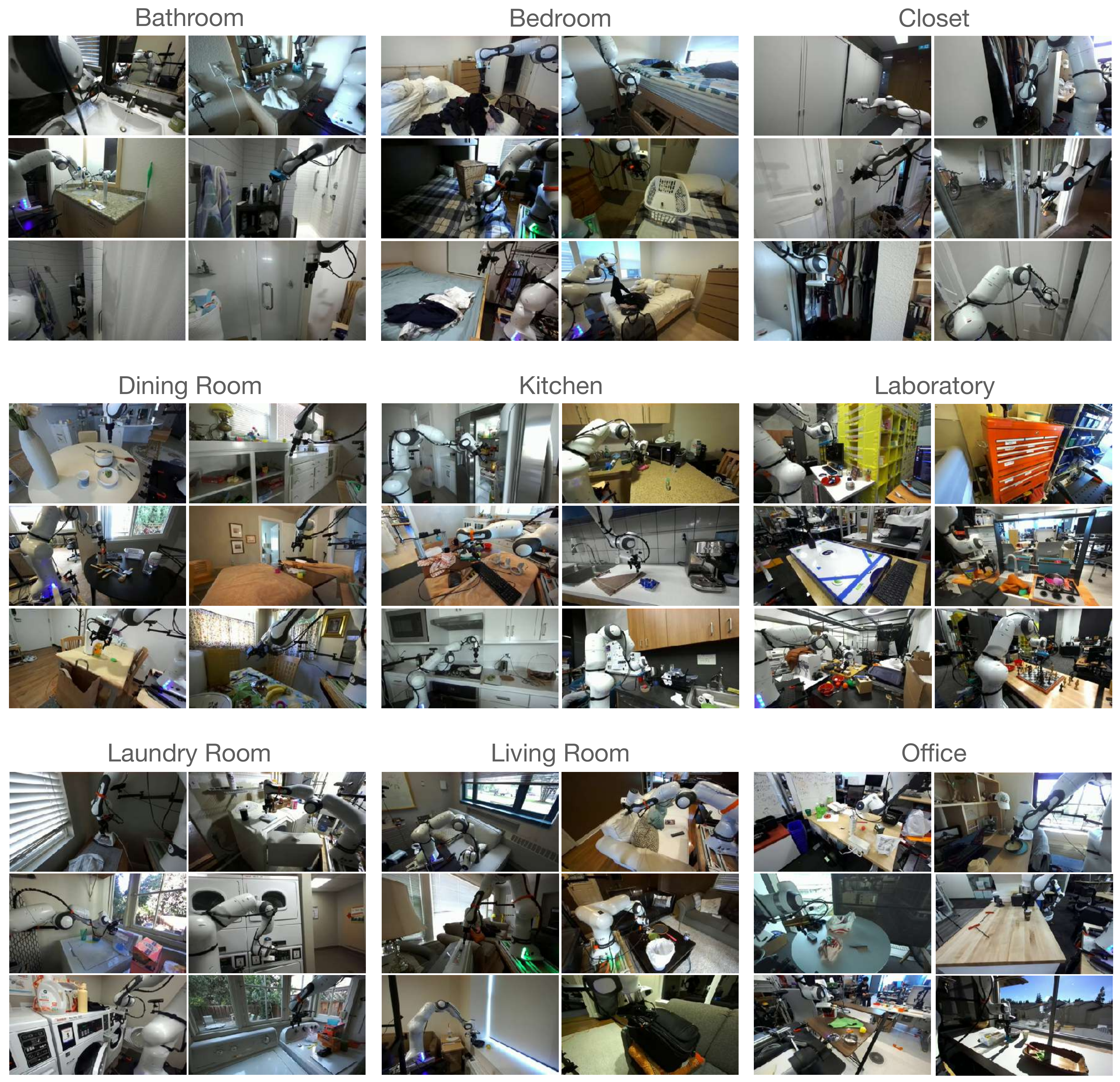}
  \caption{Qualitative examples of scenes in \name{}. We use GPT-4V to categorize scenes into 9~scene types. \name{} contains robot manipulation demonstrations in a wide range of ``in-the-wild'' scenes across \nbuildings{}~buildings. Please check out the \emph{interactive} dataset viewer included in the supplementary material to browse the dataset videos.}
  \label{fig:quali_data_examples}
\end{figure*}

\section{Scene Type Classification}
\label{sec:scene_type_classification}
We labeled scene types in an automated fashion using the GPT4V API. For each scene, we sampled a random episode from that scene and a random image from that episode. That image along with the prompt shown in Listing~\ref{lst:prompt} was sent for labeling. We then reviewed samples assigned "Other" to confirm that we were not missing any major categories, and then reassigned those labels manually.

\begin{listing*}[t]
  \centering
\begin{mdframed}[style=codebox]
\begin{verbatim}
Please classify the image into one of the following categories.
Respond with just the category name (do not include the category number).
1. Industrial office: industrial office tables and chairs, conference rooms,
conference TVs
2. Industrial kitchen: industrial refrigerator, sink, coffee maker
3. Industrial dining room: industrial setting with dining tables
4. Home office: desk or desk chairs in a home setting
5. Home kitchen: refrigerator, kitchen sink, kitchen tabletop in a home setting
6. Home dining room: dining table, dining chairs, in a home setting
7. Bedroom: room with a bed
8. Bathroom: Showers, baths, toilets, bathroom sinks
9. Living room: places with couches, armchairs, coffee tables, tvs in a home
setting
10. Hallway / closet: areas between rooms, situations where the robot is
interacting with a door or objects in a closet
11. Other: any other location that does not fit into those categories
12: Unknown: a scene that's too hard to classify because the image is dark or too
close up
\end{verbatim}
\end{mdframed}
\caption{The prompt provided to GPT4V in order to classify scene types.}
\label{lst:prompt}
\end{listing*}

\section{Indentifying Unique Scenes}
\label{sec:unique_scenes}
As mentioned in the main body of the paper, we define a unique scene as a substantial change to the robot's workspace. For example, a home kitchen may have multiple unique scenes associated with it in which the robot is placed in front of the refrigerator, sink, stove top, or different sections of the counter. We do not consider changes in the objects being interacted with or changes to the poses of external cameras sufficient to constitute a unique scene.

We label unique scenes as follows. During data collection, a scene ID number is generated each time the user indicates that robot or external cameras are moved. In total, there are 2,080 unique scene IDs in the dataset. Many of these scene IDs correspond to the same scene based on the definition provided above since users mislabel scene changes or move the robot back to the same scene after moving it somewhere else. 

In order to identify these duplicates, we collect scenes into groups that share the same robot serial number, name of the lab collecting the data, and building name. Within each group, we order the scenes by timestamp. We then go through the scenes sequentially, identifying cases where the scene did not change sufficiently to constitute a unique scene. Finally, we search across the remaining set of scenes within each group to identify cases where a robot was placed at the same scene twice (though not sequentially), and also remove these from the set of unique scenes.

This labeling approach has some limitations. For example, because we group scenes based on robot serial number and only identify duplicates within that group, if two different robots are placed at the same scene then that scene would be counted twice. Nevertheless, during labeling we were conservative in our estimate of what constituted a unique scene, and as a result believe that the number reported in the paper represents a conservative estimate.

\section{Evaluation Procedure}
\label{sec:eval_procedure}
We evaluate learned policies on the following \nevaltasks{}~tasks, each with their own out of distribution variants. For each evaluation, we ensure that each of the policies see a similar initial distribution of object locations across trials.

\noindent \textbf{Place Chips on Plate}: A short horizon task in a lab setting, where the task is to pick and place a bag of Doritos chips onto a plate, with two distractor objects on the table. All objects and the plate position are randomized between episodes on the table. We collect 50 demonstrations, and mark success if the chips are in the plate. We also consider two out of distribution variants: (1) changing the type of chips to Sun Chips (different size and color) and (2) putting two additional distractor objects (an apple and an orange) on the table. 

\noindent \textbf{Put Apple in Pot}: A medium horizon task in a lab setting, where the task is to pick and place an apple into a pot and then put a lid on the pot. The apple, pot, and lid position are randomized between episodes on the table. We collect 60 demonstrations, and mark success if the apple is in the pot and the lid is on the pot. The out of distribution variant involves placing an additional plate on the table as a distractor. 

\noindent \textbf{Toasting}: A medium horizon task in a lab setting, where the task is to put an object on a toaster oven tray, then close the toaster oven. The object and toaster position are randomized between episodes on the table. We collect 150 demonstrations, and mark success if the object is in the toaster oven and the toaster oven is closed. The out of distribution variant consists of considering novel objects to toast. 

\noindent \textbf{Closing Waffle Maker}: A short horizon task in a lab setting, where the task is to close a waffle maker. The waffle maker position is randomized between episodes. We collect 70 demonstrations, and mark success if the waffle maker is closed. The out of distribution variant consists of adding several distractor objects on the table.  

\noindent \textbf{Clean up Desk}: A long horizon task in an office setting, where the task is to open a drawer, pick and place an eraser into the drawer, and then close the drawer. The eraser position is varied at the start of each episode at a set schedule of different positions and orientations. We collect 50 demonstrations, and mark success if the drawer is closed with the eraser in it. The out of distribution variant consists of adding distractor objects on the desk, specifically a calculator, three whiteboard markers, and a clip. We found that adding distractor objects inside the desk caused all policies to fail.

\noindent \textbf{Cook Lentils}: A long horizon task in a kitchen setting, where the task is to remove the lid off a pan, pour lentils into the pan, and turn on the stove. The object positions are fixed. We collect 50 demonstrations, and mark success if all 3 stages of the task are successfully completed. The out of distribution variant consists of adding several distractor objects and a camera shift.

\section{Diffusion Policy Details}
\label{sec:policy_details}
In Section~\ref{subsec:policy_arch} we discuss the policy architecture and hyperparameters used for all policy learning experiments. Then in Section~\ref{subsec:batch_construction}, we describe how the various datasets in the paper are used to construct training batches for policy learning.

\subsection{Diffusion Policy Architecture and Hyperparameters}
\label{subsec:policy_arch}

We build our diffusion policy~\citep{chi2023diffusionpolicy} training pipeline on the Robomimic codebase~\citep{robomimic2021}, which provides high quality implementations of a number of different imitation learning and offline RL algorithms. Given camera observations and a language instruction for the task, within Robomimic, we define observation keys corresponding to each of the two external camera observations, a frozen DistilBERT~\citep{Sanh2019DistilBERTAD} language embedding, the 3D cartesian position of the gripper, and the gripper state, which measures the degree to which the gripper is closed. 

For each of the camera observations, we first downsample each image to a resolution of $128 \times 128$ and apply color jitter and random cropping as a form of data augmentation. We then use a ResNet-50 visual encoder pre-trained on ImageNet~\citep{deng2009imagenet} to produce embeddings for each of the visual inputs. These embeddings are directly concatenated with all of the other observation keys. These concatenated features are then fed through an Observation Processing MLP with layers defined in Table~\ref{tab:hyperparameters}. The output of this MLP is then passed to a U-Net diffusion head which generates action trajectories. We use an observation horizon of 2 to condition the diffusion head, diffuse out 
16-step action sequences, and step the first 8 actions open loop before re-running policy inference. All relevant hyperparameters are defined in Table~\ref{tab:hyperparameters}. In line with prior work~\citep{chi2023diffusionpolicy}, we use DDIM to diffuse out action trajectories for improved efficiency.

All experiments use the training hyperparameters in Table~\ref{tab:hyperparameters} with one exception: for the Cook Lentils task OOD experiment, we train all policies with $50000$ training steps due to the increased complexity of the task.

\begin{table}[hbtp]
    \centering
    \caption{\textbf{Training Hyperparameters}}
    \label{tab:hyperparameters}
    \begin{tabular}{@{}cc@{}}
    \toprule
    \textbf{Hyperparameter}    & \textbf{Value}                         \\ \midrule
    Batch Size              & 128                           \\
    Optimizer               & Adam \\ 
    Learning Rate           & 1e-4 \\
    Learning Rate Scheduler & Linear \\
    Train Steps & 25000 \\ \bottomrule%
    Observation Processing MLP & [1024, 512, 512] \\
    Image Resolution & (128, 128) \\
    Crop Height & 116\\
    Crop Width & 116\\ \\ \bottomrule
    Diffusion Method & DDIM \\ 
    EMA Power & 0.75 \\
    U-Net Hidden Layer Sizes & [256, 512, 1024] \\
    Observation Horizon & 2 \\
    Prediction Horizon & 16 \\
    Action Horizon & 8 \\ \bottomrule
    \end{tabular}
\end{table}

\begin{figure*}[!htbp]
  \centering
  \includegraphics[width=1.0\linewidth]{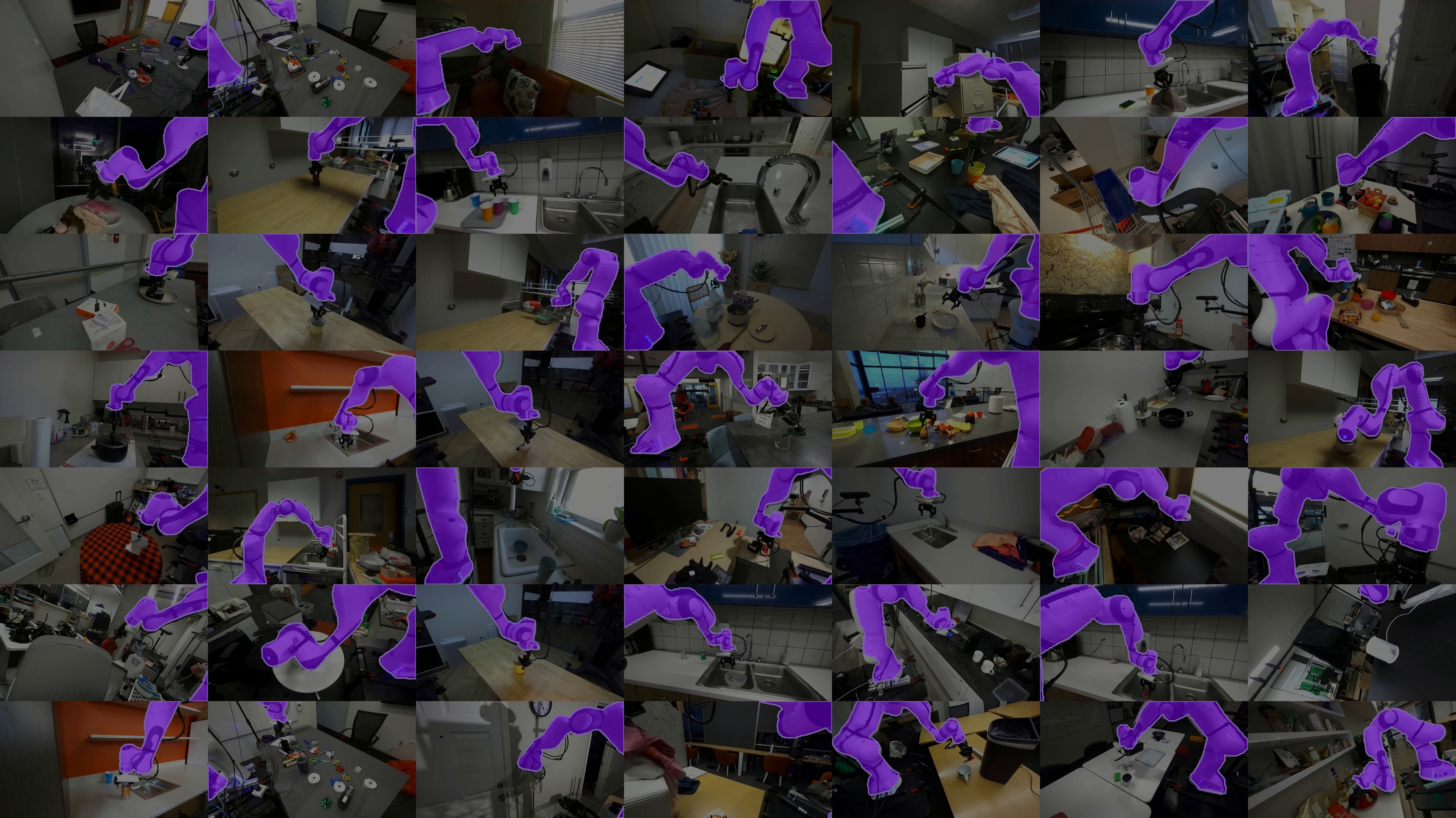}
  
  \caption{
  \textbf{Camera-to-robot base calibration qualitative results} showing randomly picked scenes with synthetically rendered robot masks using PyTorch3D~\citep{ravi2020pytorch3d}. Renderings are generated by importing the robot’s URDF-defined mesh and kinematic structure, applying joint angles to compute the articulated pose, and transforming the mesh to the camera frame using the extrinsic $\textbf{T}_{\text{cam} \rightarrow \text{base}}$. The extrinsic results are a combination of results from automatic quality assessment-based filtering, outlined in Sec.~\ref{subsec:existing_cam2base_quality} and running a tuned CtRNet-X model~\citep{lu2024ctrnet}, outlined in Sec.~\ref{subsec:auto_cam2base_quality}. We provide quality assessment metrics for both approaches in our released extrinsic.
  } 
  \label{fig:cam2base_calib}
\end{figure*}

\subsection{Training Batch Construction}
\label{subsec:batch_construction}
For each evaluation task, we train policies with 3 different methods of constructing training batches:
\begin{itemize}
    \item \textbf{No Co-training}: trains a state of the art diffusion policy~\citep{chi2023diffusionpolicy} using samples from the in-domain demonstrations only.
    \item \textbf{\name{} (Ours)}: Trains a diffusion policy, but mixes batches 50/50 between in-domain demonstrations and \name{} trajectories. For this experiment, we consider the first 40K successful trajectories in \name{} for which language annotations were available at the time of policy training. For the \textit{scene diversity} experiments, we use a 7362 trajectory subset of these 40K trajectories.
    \item \textbf{OXE~\citep{open_x_embodiment_rt_x_2023}}: Trains a diffusion policy, but mixes batches 50/50 between in-domain demonstrations and trajectories from a curated subset of the Open X-Embodiment dataset~\citep{open_x_embodiment_rt_x_2023} (OXE) used in \citet{octo_2023}. We also omitted data from the language table split of OXE to bring down the number of trajectories to a manageable scale (400K trajectories).
\end{itemize}

Each of the above settings defer only in the data used to construct each training batch: otherwise all policies have identical architectures and are trained with the same training parameters specified in Section~\ref{subsec:policy_arch}.

The in-domain demonstrations used for policy training consist of only the demonstrations collected for each evaluation task in Section~\ref{sec:eval_procedure} with one exception: for the Toasting and Close Waffle Maker Tasks, one multi-task policy is trained on the combination of their demonstrations. Thus, in this case, the \textbf{No Co-training} policy defined above trains one diffusion policy on the \textit{combined} in-domain demonstrations, while the Co-training experiments sample batches via a 50/50 split of data from these combined in-domain demonstrations and data from either \name{} or \textbf{OXE~\citep{open_x_embodiment_rt_x_2023}}.

\section{Automatic Camera Calibration}
\label{subsec:auto_camera_calib}

\begin{figure*}[!htbp]
  \centering
  \includegraphics[width=1.0\linewidth]{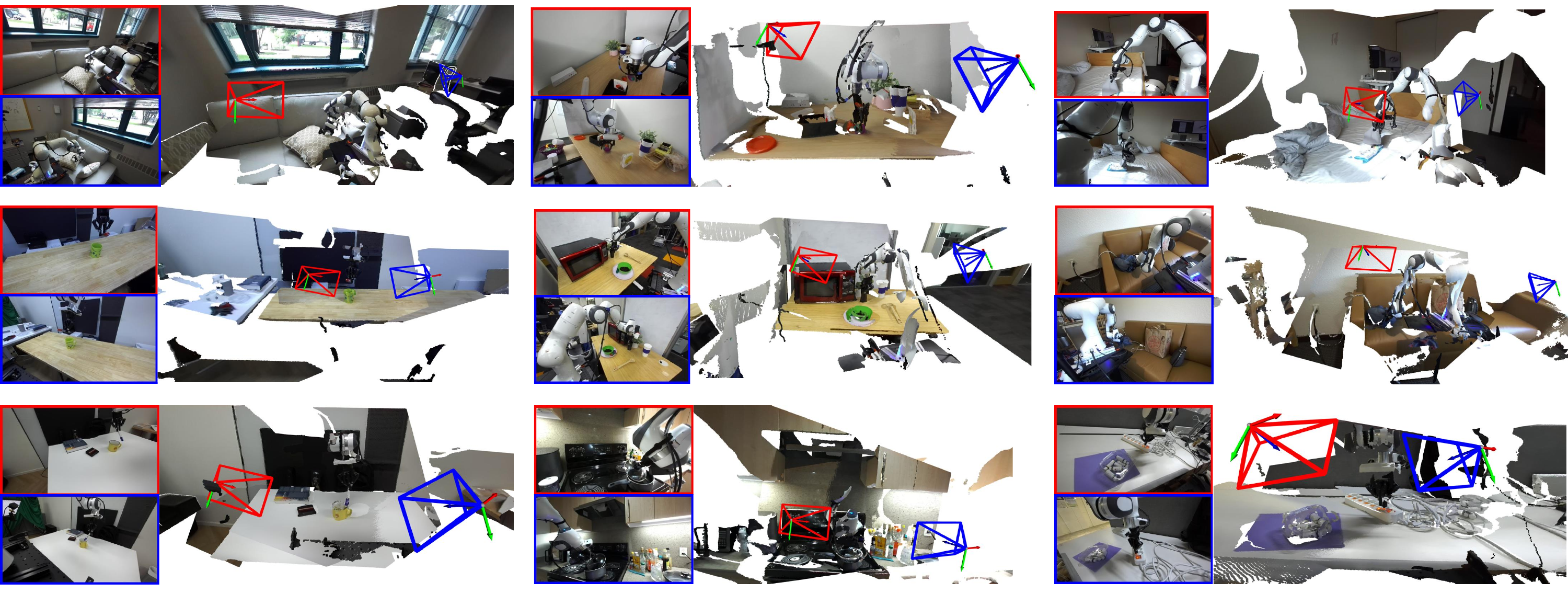}
  
  \caption{
  \textbf{Camera-to-Camera calibration qualitative results} showing images, camera poses and pointclouds after our improved off-line and post-hoc camera calibration as discussed in Sec.~\ref{subsec:auto_cam2_calib}. Scenes are picked from the top 30\% quantile based on the number of matches after calibration~(See. Figure~\ref{fig:cam2cam_plot}) External cameras are shown in red and blue. Here accumulated pointclouds from both views are shown after deprojecting the depth maps using camera intrinsics and accumulated using relative camera poses between the two cameras. 
  } 
  \label{fig:cam2cam_calib}
\end{figure*}

In this section, we provide \textbf{three comprehensive sets of camera calibration matrices} for the DROID dataset with their \textbf{respective quality assessment metrics}, including \textbf{camera-to-base calibrations for 36k unique scenes} with one of the cameras calibrated with respect to base, \textbf{camera-to-camera calibrations for all scenes}, and a curated \textbf{superset of 24k scenes} encompassing all three methods and with both cameras calibrated with respect to base, facilitating downstream robust geometric understanding in robotics and 3D perception tasks.

Accurate camera calibration can be very useful in robotics and 3D perception, as it enables the consistent encoding of spatial geometry from visual data. It serves as the backbone for various downstream tasks in robotics manipulation, such as learning viewpoint invariant representations~\citep{chen2024roviaug,tian2024vista} or grounding actions through 3D vision-and-language models~\citep{shridhar2023perceiver, zhen20243d}, thus enabling the robotics agents to achieve geometric and visual generalization. In robotic applications, calibration allows for precise scene understanding and interaction by aligning sensor observations to a shared spatial frame.

The DROID dataset provides initial extrinsic parameters~(Sec.~\ref{sec:dataset}) that transform coordinates from the camera frame to the robot base frame. However, these calibrations are not always accurate, primarily due to slight errors that can arise during the manual calibration process, such as imperfect checkerboard placements, variations in lighting conditions, or inaccuracies in the OpenCV calibration procedure performed at the start of each data collection session.
Following the data collection efforts outlined in Sec.~\ref{sec:dataset}, we additionally focus on providing robust calibration values for the collected dataset in an off-line post-hoc manner. This process utilizes recent advances in deep-learning based perception systems~\citep{kirillovSegmentAnything2023, wang2024dust3rgeometric3dvision, radford2021learningtransferablevisualmodels, lu2024ctrnet} to automatically calibrate the relevant cameras and provide quality metrics in a post-hoc manner. %

The following sections details the automatic post-hoc calibration of the pre-collected DROID dataset. It focuses on two key types of calibration: (i) camera-to-robot base extrinsic calibration, which computes the transformation between a fixed camera and the robot’s kinematic base; and (ii) camera-to-camera extrinsic calibration, which estimates the relative pose i.e. orientation and translation between two external cameras. Both are essential for fusing multi-view observations as well as allowing the grounding of robotics actions in 3D, thus enabling spatially grounded robotic behaviors. This section is divided into 3 sub-sections. We first detail the quality metric assessment
of existing camera-to-robot base calibration~(Sec.~\ref{subsec:existing_cam2base_quality}) provided after the data-collect phase in Sec.~\ref{sec:dataset}. This would allow us to filter the extrinsics provided during data-collect and provide certain guarantees regarding the calibration already provided. We then explain how to calibrate additional cameras with respect to the base using a fully automatic pipeline~\citep{lu2024ctrnet} while also providing guarantees in terms of quality metric i.e. reprojection error~(Sec.~\ref{subsec:auto_cam2base_quality}). Furthermore, we discuss calibrating external cameras with respect to each other in Sec.~\ref{subsec:auto_cam2_calib}. Finally, we include a discussion on limitations and future work in Sec.~\ref{subsec:limitation_and_future_work}.

\subsection{Quality Assessment of Existing Camera-to-Robot Base Calibration}
\label{subsec:existing_cam2base_quality}

To evaluate the quality of the existing camera-to-robot base transformation ($\mathbf{T}_{\text{cam} \rightarrow \text{base}}$), we project known 3D keypoints $\mathbf{X} \in \mathbb{R}^{N \times 3}$, obtained via forward kinematics for given joint angles $\theta$, into the image plane using the extrinsic matrix $\mathbf{T}_{\text{cam} \rightarrow \text{base}}$ and camera intrinsics $\mathbf{K}$. The 2D projections $\mathbf{x} \in \mathbb{R}^{N \times 2}$ are computed as $\mathbf{x} = \pi(\mathbf{K} \cdot \mathbf{T}_{\text{cam} \rightarrow \text{base}} \cdot \mathbf{X})$, where $\pi(\cdot)$ denotes perspective projection followed by normalization.

These 2D keypoints are used to guide a Segment-Anything~(SAM)~\citep{kirillov2023segment} instance segmentation model, which predicts masks $\mathcal{M}_{\text{SAM}}$. Simultaneously, synthetic robot masks $\mathcal{M}_{\text{GT}}$ are rendered using PyTorch3D by importing the robot’s mesh geometry and kinematic structure defined in its URDF. Each joint angle configuration $\theta$ is applied to the URDF to compute the articulated 3D mesh pose of the robot. The resulting mesh is transformed to the camera frame using the same extrinsic transformation $\mathbf{T}_{\text{cam} \rightarrow \text{base}}$. The posed mesh is then rasterized into a binary silhouette using a differentiable renderer with the corresponding camera intrinsics $\mathbf{K}$. This rendered mask serves as the ground-truth projection for evaluating the alignment quality of the predicted segmentation.

We compute the Intersection-over-Union (IoU) between the predicted and ground-truth masks as $\text{IoU} = \frac{|\mathcal{M}_{\text{SAM}} \cap \mathcal{M}_{\text{GT}}|}{|\mathcal{M}_{\text{SAM}} \cup \mathcal{M}_{\text{GT}}|}$. Only SAM masks with confidence scores greater than 0.65 are retained. A final threshold of $\text{IoU} \geq 0.7$ is used to identify high-quality projections, filtering out poorly aligned frames. We report the mean IoU across 5 equally subsampled frames in a video sequence as a measure of calibration quality. Using this process, we identified a total of around 30k scenes with either the left or right camera well calibrated with respect to the scene. The whole process took around 1 day on 8-A100 Nvidia-GPUs. 

\begin{figure}[!b]
  \centering
  \includegraphics[width=1.0\linewidth]{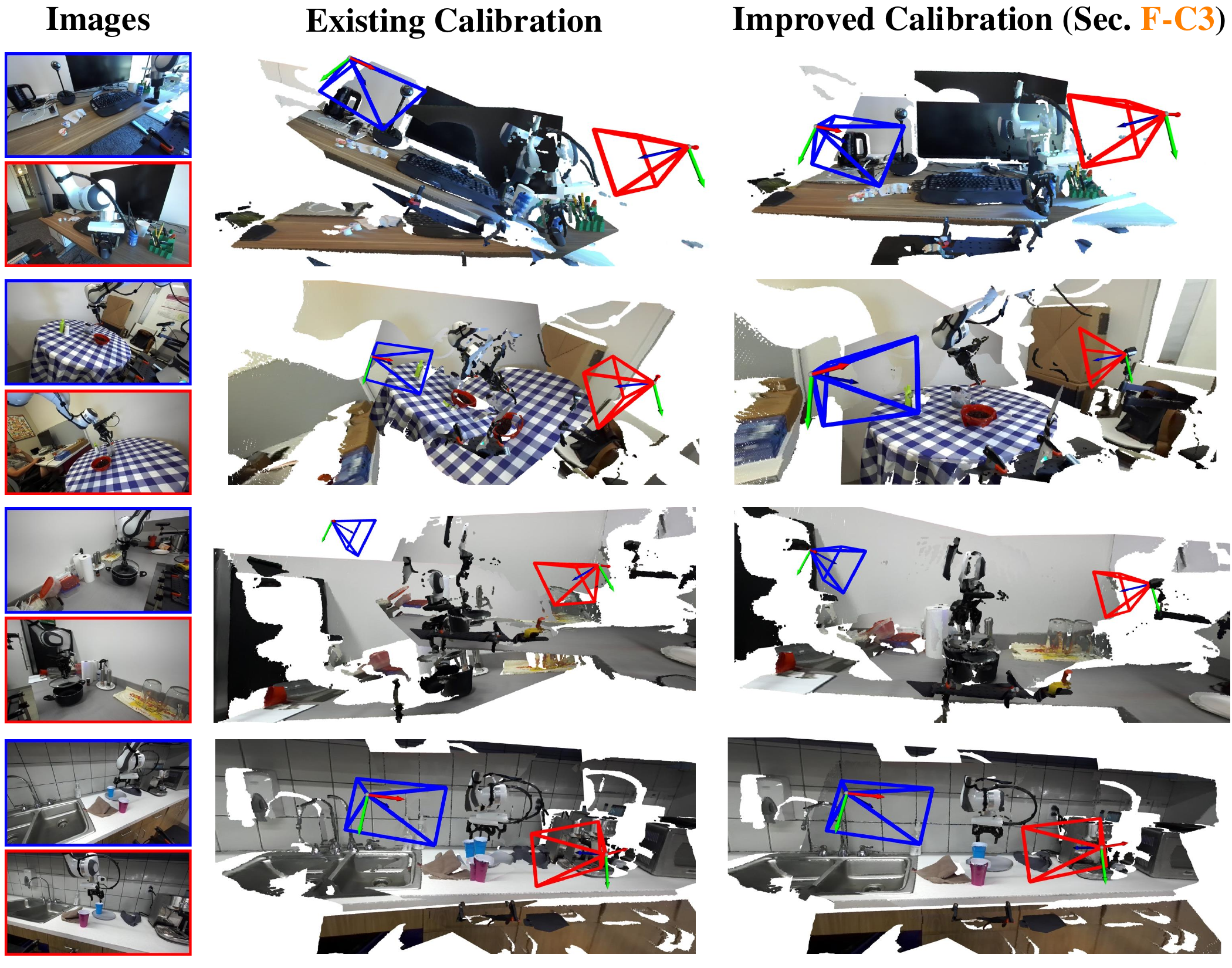}
  \caption{
    \textbf{Camera-to-Camera calibration comparison} showing images~(\emph{left}), pointclouds from the existing calibration~(\emph{middle}) and pointclouds after our improved calibration~(\emph{right}), as described in Sec.~\ref{subsec:auto_cam2_calib}. Our improved calibration is able to handle challenging scenes and produces well-aligned pointclouds from both cameras. Note that depth maps which are used to deproject pointclouds using camera intrinsic and extrinsic are not shown here.
  } 
  \label{fig:cam2cam_calib_2}
\end{figure}

\begin{figure*}[!t]
  \centering
  \includegraphics[width=1.0\linewidth]{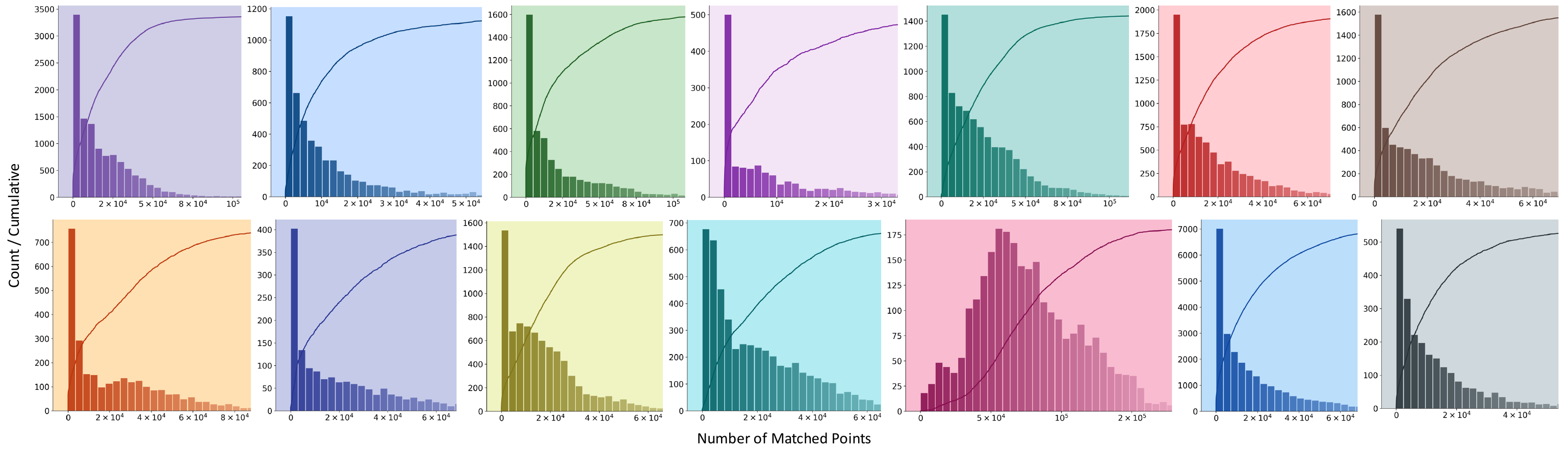}
  
  \caption{
    \textbf{Distribution of matched points} after camera-to-camera calibration for each unique lab along with the cumulative distribution. While some labs achieve high-quality correspondence, others struggle to reach the same level — often due to challenging lighting or clutter. The cumulative curve (solid curve line) highlights the accumulation of matched points across all scenes, helping to identify the top quantile of well-calibrated camera pairs within each lab. These high-confidence matches are especially important as they inform downstream selection of reliable scenes. Note that the first image from each video was used for pose refinement using the modified DUSt3R~\citep{wang2024dust3rgeometric3dvision} pipeline described in Sec.~\ref{subsec:auto_cam2_calib}.
  } 
  \label{fig:cam2cam_plot}
\end{figure*}

\subsection{Automatic Camera-to-Robot Base Calibration}
\label{subsec:auto_cam2base_quality}

To supplement the filtering strategy outlined in Sec.~\ref{subsec:existing_cam2base_quality} and bring in additional cameras for the camera-to-robot base calibration, we additionally ran a tuned version of CtRNet-X~\citep{lu2024ctrnet} out-of-the-box on all of DROID dataset. We used the original codebase provided by the authors as well as the hyperparameters tuned for the DROID dataset. CtRNet-X is a feed-forward approach that detects keypoints on robot using a neural-network and matches it with ground-truth 3D keypoint trajectory in the video. Additionally, they utilize a CLIP~\citep{radford2021learningtransferablevisualmodels}-guided robot part detection to dynamically select visible keypoints. Following author's implementation, we use a confidence threshold of 0.08, CLIP~\citep{radford2021learningtransferablevisualmodels} models' end-effector confidence of 0.1 and robot base confidence of 0.05. To evaluate the quality of our camera-to-base calibration, we compute the reprojection error between detected 2D keypoints and their corresponding 3D projections using the estimated camera pose and intrinsics. To ensure robustness, we first discard low-confidence 2D observations based on a fixed threshold. We then apply a Median Absolute Deviation (MAD) based outlier rejection strategy: we calculate the median of all reprojection errors, compute the absolute deviation of each error from the median, and identify inliers as those within 2.5 times the MAD. This robust statistical filtering helps suppress the influence of large outliers, leading to a more reliable estimate of the mean reprojection error. For the final filtering, we select scenes with a mean reprojection-error of 20 or less. 

\begin{figure}[!t]
  \centering
  \includegraphics[width=1.0\linewidth]{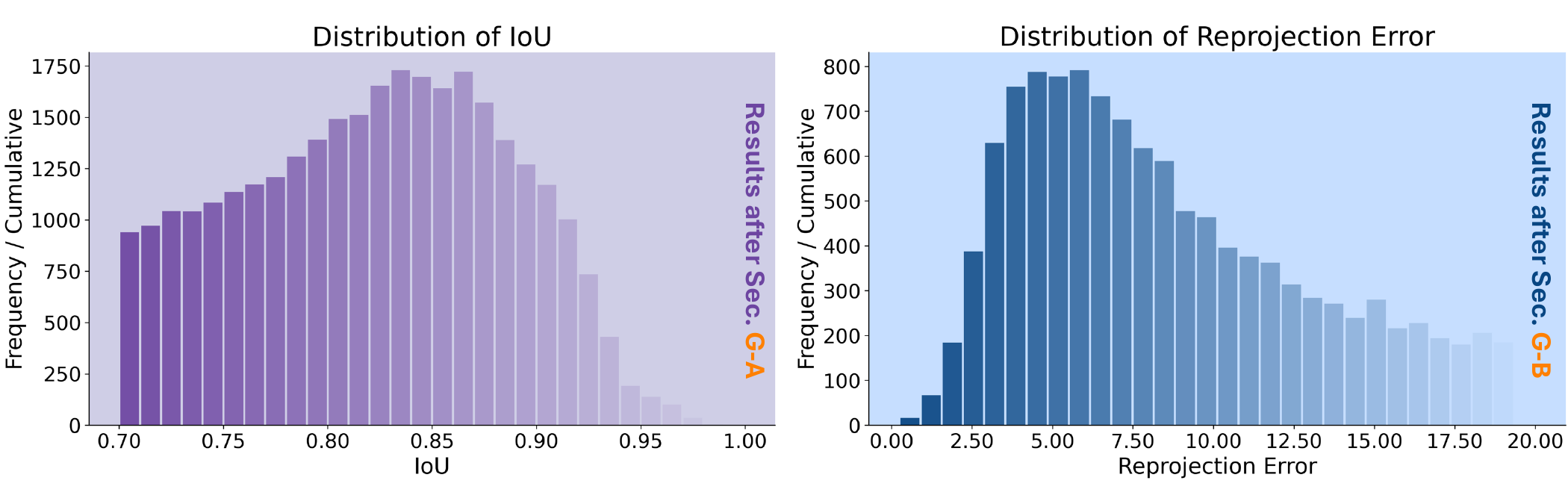}
  
  \caption{
    \textbf{Distribution of respective metrics} i.e. IOU and mean reprojection errors after thresholding and filtering with the strategy outlined in Sec.~\ref{subsec:existing_cam2base_quality} and Sec.~\ref{subsec:auto_cam2base_quality} respectively.
  } 
  \label{fig:cam2base_histograms}
\end{figure}

Through this process, we identified a total of around 12k scenes which have either the left or right camera correctly calibrated with respect to the base. This process took around 5 days on 8-A100 Nvidia GPUs. Since the number of well-calibrated scenes overlapped between the two strategies, we are able to calibrate around 36k unique scenes with either the left or right camera well calibrated with respect to the scene using the strategy outlined in Sec.~\ref{subsec:existing_cam2base_quality} and Sec.~\ref{subsec:auto_camera_calib}. Figure~\ref{fig:cam2base_calib} shows randomly selected scenes with synthetically rendered robot masks using PyTorch3D~\citep{ravi2020pytorch3d}, qualitatively demonstrating the high accuracy of our filtered camera-to-robot base calibration from both the above-mentioned approaches. Furthermore, we present the distribution of IoU and reprojection errors after applying the improved calibration strategy and filtering. These results are shown in Fig.~\ref{fig:cam2base_histograms}.

\subsection{Automatic Camera-to-Camera Calibration}
\label{subsec:auto_cam2_calib}

We utilize the recently released DUSt3R~\citep{wang2024dust3rgeometric3dvision} framework for improved Camera-to-Camera calibration. DUSt3R~\citep{wang2024dust3rgeometric3dvision} supports both relative and absolute pose estimation. For relative pose estimation, DUST3R proposes obtaining 2D--3D correspondences between a query image $I_Q$ and a reference image $I_B$, followed by PnP-RANSAC~\citep{fischler1981random, lepetit2009epnp} using known or estimated intrinsics. The relative pose between $I_Q$ and $I_B$ can also be converted to an absolute pose in world coordinates by aligning predicted pointmaps to a known scale, typically via a ground truth pointmap for $I_B$. However, this approach still requires scale alignment post-optimization and can suffer from ambiguities due to noise or uncertainty in the predicted geometry.

We modify the pose optimization pipeline of DUSt3R~\citep{wang2024dust3rgeometric3dvision} to utilize depth maps and known camera intrinsics as an input in the optimization pipeline to recover  absolute poses in a consistent metric scale. Specifically, we begin by running DUSt3R inference on image pairs to extract dense 3D pointmaps. These predicted pointmaps are aligned to ground truth 3D point clouds—constructed from depth and intrinsics—to compute a global scale factor. We then perform a global optimization step, where the ground truth depth and intrinsics are fixed, and camera poses are refined to minimize the 3D alignment error across the scene. This approach enables accurate, scale-aware absolute pose estimation without relying on post-hoc scale alignment.

By fixing the depth and intrinsics during optimization, we ensure that the recovered poses are globally consistent and metrically accurate. Importantly, our method operates directly on unmodified DUSt3R~\citep{wang2024dust3rgeometric3dvision} outputs, requiring no additional training or manual scale correction. Figures~\ref{fig:cam2cam_calib} and~\ref{fig:cam2cam_calib_2} show qualitative improvements in point cloud alignment following this optimization step.

To assess the quality of the recovered camera-to-camera calibration, we report the number of matched points between views, following the original implementation by \citep{wang2024dust3rgeometric3dvision}. For each image pair, we extract high-confidence 3D points, project them into 2D using known intrinsics, and identify reciprocal nearest neighbors in 3D space as reliable matches. Formally, given pointmaps $P_0$ and $P_1$, we define the match set $\mathcal{M} = \left\{ (i, j) \mid \mathrm{NN}(P_0[i]) = j \text{ and } \mathrm{NN}(P_1[j]) = i \right\}$. In practice, we qualitatively observe that a higher number of reciprocal 3D matches visually correlates with the geometric quality of estimated poses~(see Fig.~\ref{fig:cam2cam_calib}). Although the number of matches serves as a reasonable proxy for assessing the quality of estimated poses, we observed some false positives in visually cluttered scenes. To enhance robustness, one could lower the filtering threshold or incorporate the quality assessment described in Sec.~\ref{subsec:existing_cam2base_quality} to further refine the filtering process. Figure~\ref{fig:cam2cam_plot} shows the distribution of match counts across labs. While some labs exhibit strong geometric consistency, others struggle due to challenging conditions like clutter or poor lighting. For all videos, the first frame was used for pose refinement via our modified DUSt3R~\citep{wang2024dust3rgeometric3dvision} pipeline. Future improvements could include ensembling predictions across frames or leveraging temporal consistency to further stabilize pose estimation or finetuning DUST3R-like methods on table-top cluttered datasets observed in robotics manipulation settings.

\subsection{Limitations and Future Work}
\label{subsec:limitation_and_future_work}

Calibrating a large-scale dataset like DROID is a challenging task. To ensure accuracy and provide guarantees at each step, we divided the calibration process into three distinct stages. Since the complete process is fully automatic, there are still some false positives and future work could look at further improving on these inconsistencies. Part of our camera-to-base calibration relied on running an out-of-the-box model which was trained on Franka panda robot. While successful, its zero-shot generalizability to other robots (without requiring further training or finetuning) remains to be seen. 

Future work could look at using foundation models to segment out the robot or gripper and estimating keypoints on specific parts of the robot. This could provide a more generalizable solution that could be readily applied to any robot collected data in-the-wild. Our Camera-to-Camera calibration relied on a recent paradigm in 3D deep learning, namely the prediction of point-maps. Despite using a modified version of DUSt3R~\citep{wang2024dust3rgeometric3dvision}, which utilized privileged depth information for pose optimization, it relied on the original checkpoints provided by the authors and hence also borrowed the limitations of the original model. Despite decent success, the model at times fails on scenes with clutter and challenging table-top settings with little to no overlap between images. As the quality of these models keeps improving~\citep{Yang_2025_Fast3R, wang2025vggt}, we believe it would be a valuable direction to leverage these improved models for more robust camera-to-camera calibration, particularly in cluttered and low-overlap scenarios where traditional feature matching or earlier models struggle. 

\subsection{Conclusion}
The approaches outlined in each of the aforementioned sections i.e. Sec.~\ref{subsec:existing_cam2base_quality}, Sec.~\ref{subsec:auto_cam2base_quality} and Sec.~\ref{subsec:auto_cam2_calib} have different guarantees in terms of quality metrics. Hence, for the final calibration release we offer 3 different sets of camera calibration matrices. The first one contains Camera-to-Robot base calibration from 36k unique scenes with either the left or right camera calibrated with respect to the robot base. This includes results after the combined calibration methods outlined in Sec.~\ref{subsec:existing_cam2base_quality} and Sec.~\ref{subsec:auto_cam2base_quality}. The second set of calibration includes all Camera-to-Camera calibration matrices, i.e. the relative transformations for all scenes in DROID dataset using the approach outlined in Sec.~\ref{subsec:auto_cam2_calib}. Finally, we also release a third set of calibration with includes a superset of all methods which, totaling around 24k scenes with both cameras calibrated with respect to the base and with a mix of 3 different approaches and individual guarantees. In this superset, we use an IOU threshold of 0.6, a reprojection error of 20 and top 30\% quantile based on number of matches for each stage of calibration described earlier in Sections~\ref{subsec:existing_cam2base_quality},~\ref{subsec:auto_cam2base_quality} and~\ref{subsec:auto_cam2_calib} respectively. We hope this effort is useful for 3D vision and robotics manipulation research and also serves an inspiration for off-line automatic camera calibration of in-the-wild robotics manipulation datasets. 

\begin{figure*}[t]
  \centering\includegraphics[width=1.0\linewidth]{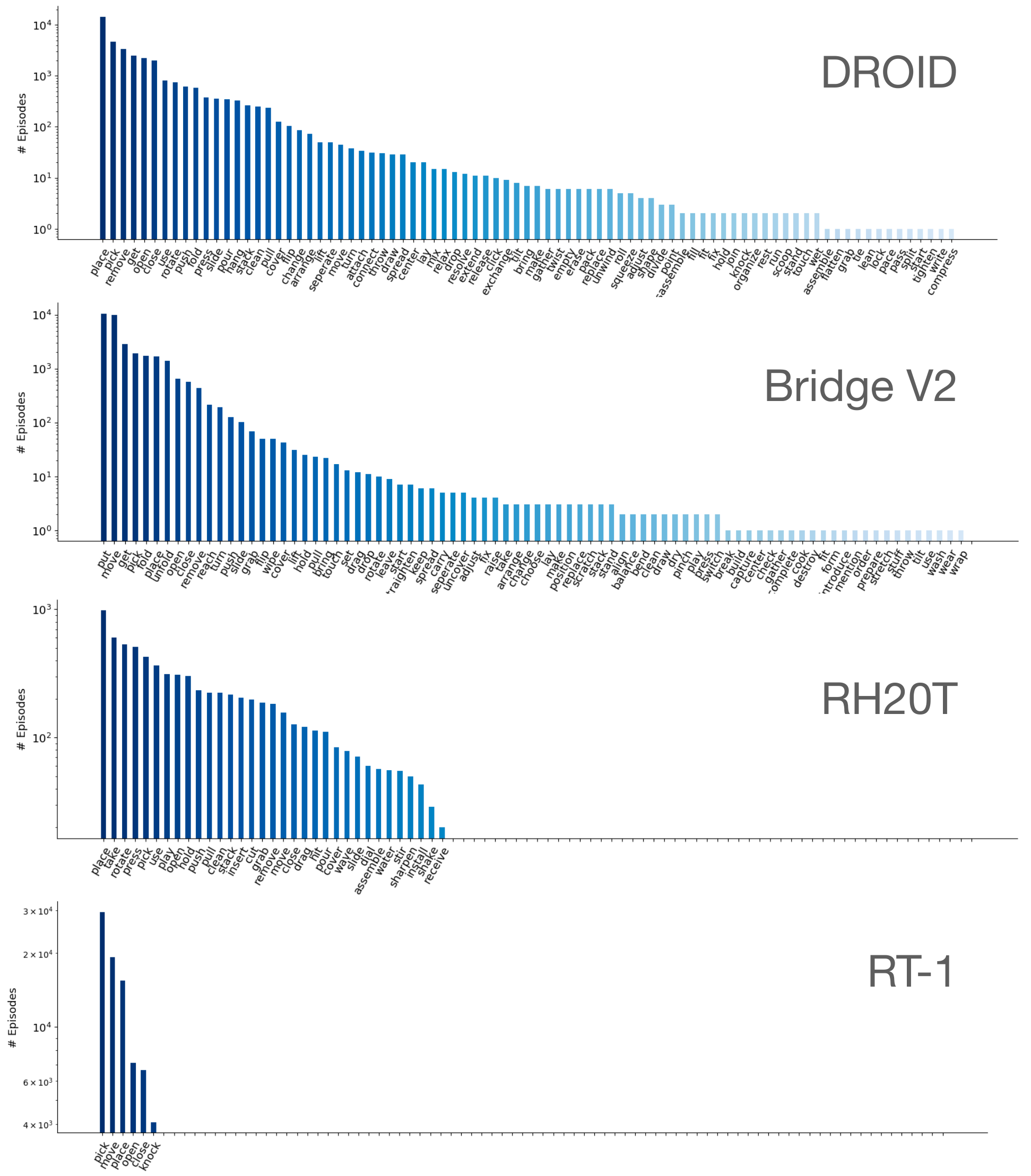}
  \caption{Distribution of skills, \ie verbs, for \name{} and existing large robot manipulation datasets. \textbf{Top to bottom}: \name{}, Bridge~V2~\citep{walke2023bridgedata}, RH20T~\citep{fang2023rh20t}, RT-1~\citep{brohan2022rt}. \name{} features a long tail of diverse verb classes that is only matched by Bridge~V2, while the RH20T and RT-1 datasets have a more constrained set of skills.}
  \label{fig:verb_distribution}
\end{figure*}

\begin{figure*}[t]
  \centering
  \includegraphics[width=1.0\linewidth]{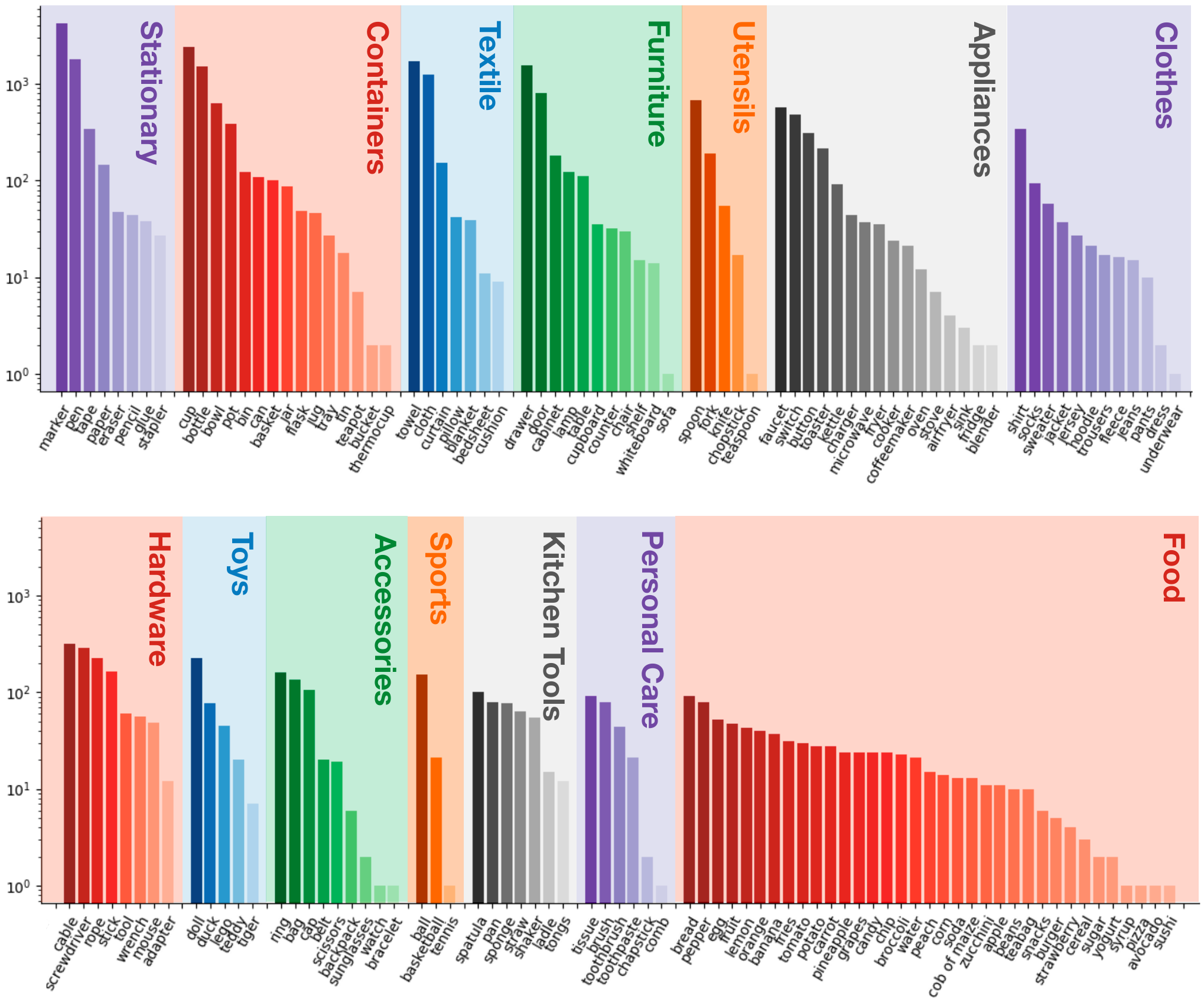}
  \caption{Distribution of interacted objects in \name{}, grouped by category. The robot interacts with a wide range of everyday objects.}
  \label{fig:object_distribution}
\end{figure*}

\begin{figure*}[t]
  \centering
  \includegraphics[width=1.0\linewidth]{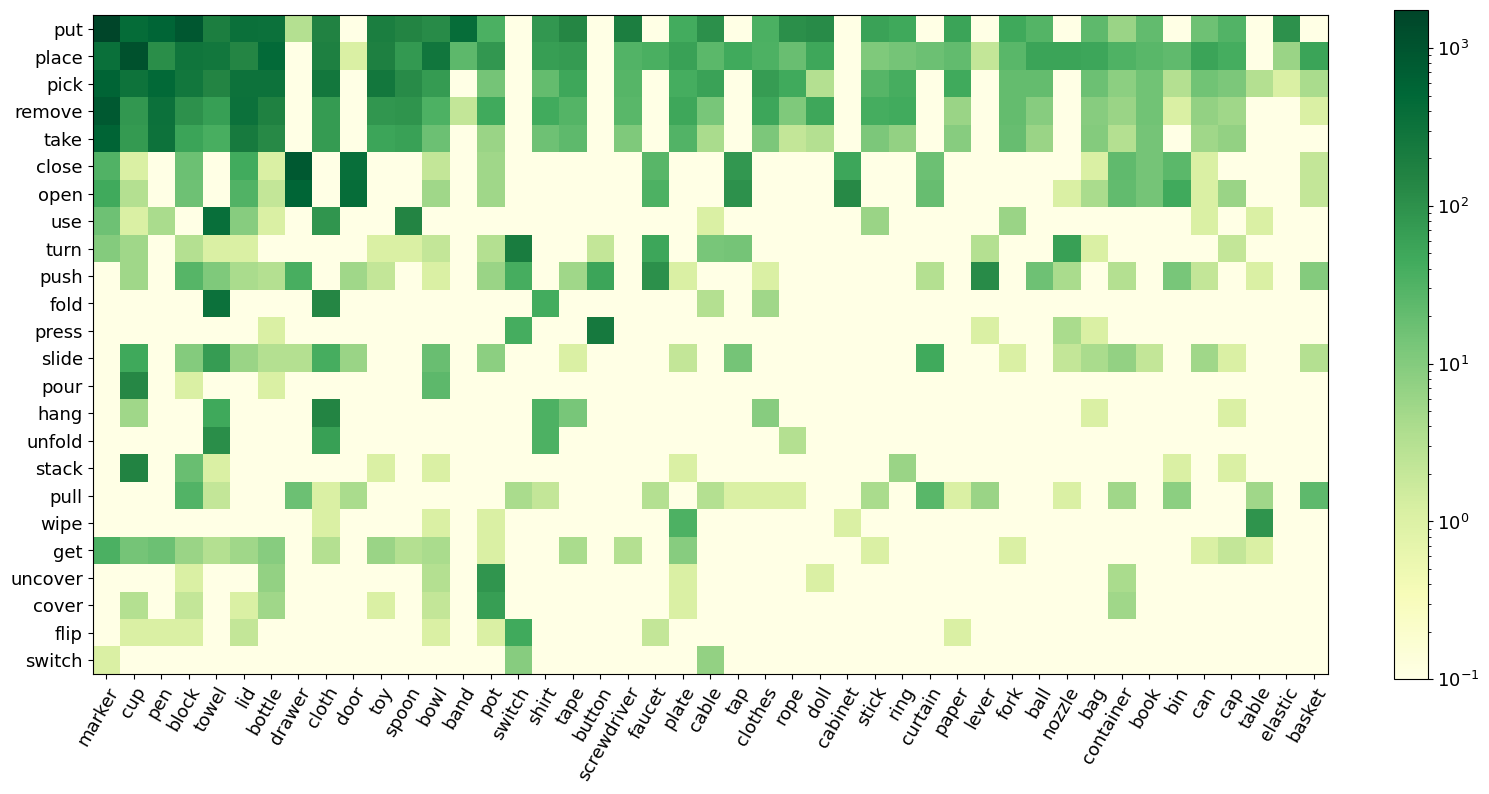}
  \caption{Joint distribution of verbs and interacted objects in \name{}. Most objects have a diverse range of interactions that are performed on them.}
  \label{fig:verb_object_heatmap}
\end{figure*}

\end{appendices}

\end{document}